\documentclass{fairmeta}
\usepackage{microtype}
\usepackage{graphicx}
\usepackage{placeins}
\usepackage{needspace}

\usepackage{booktabs,makecell,multirow,adjustbox}
\usepackage{tabularx}
\usepackage{array}
\usepackage{amsmath,amssymb}
\usepackage{pifont}
\usepackage{xcolor}
\usepackage{colortbl}
\usepackage{xurl}
\definecolor{EAgroupgray}{RGB}{242,242,242}

\title{EgoActionBench: Benchmarking Dense Action Captioning for Egocentric Human Videos}
\author[1]{Yujiao Shen}
\author[2]{Jiahe Song}
\author[1]{Haitian Li}
\author[1]{Jingkang Yang}
\author[1]{Ziwei Liu}
\affiliation[1]{S-Lab, Nanyang Technological University}
\affiliation[2]{Shanghai Jiao Tong University}
\abstract{\frenchspacing Learning from human data offers a scalable path toward embodied intelligence, yet it requires models to accurately understand fine-grained human actions from egocentric observations. However, existing benchmarks remain limited in action atomicity, temporal precision, and hand–action association. We introduce \textsc{EgoActionBench}, a benchmark that represents continuous human manipulation as temporally localized atomic actions associated with the left hand, right hand, or both hands. \textsc{EgoActionBench} contains 1,407 episodes with 41,132 human-refined action annotations.
To evaluate such dense predictions, we propose an evaluation protocol that retrieves temporally eligible candidates, selects a prediction by content agreement, and scores semantic fidelity and temporal overlap alongside action precision and recall, jointly capturing what happens, when it happens, and which hand performs it. We benchmark 20 baseline models and systems on \textsc{EgoActionBench} and further adapt two backbones, Qwen3-VL-4B and LLaVA-OneVision-2-8B, through supervised fine-tuning followed by reinforcement learning on automatically annotated training episodes. EgoAction improves EgoActionScore by 15.9 and 23.2 points on the two backbones, and both adapted models outperform all open-source baselines. Our experiments reveal substantial limitations of current models in action atomicity, temporal precision, and hand–action association, highlighting the gap between general video understanding and the action understanding required for embodied intelligence.
\par\vspace{\baselineskip}}
\metadata[Project Page]{\url{https://egoaction-bench.github.io}}
\metadata[Code]{\url{https://github.com/player0718/EgoActionBench}}
\metadata[Dataset]{\url{https://huggingface.co/datasets/play0718/EgoActionBench}}
\date{November 3, 2026}

\begin{document}

\maketitle

\section{Introduction}

\begin{figure}[t]
\centering
\includegraphics[width=\linewidth]{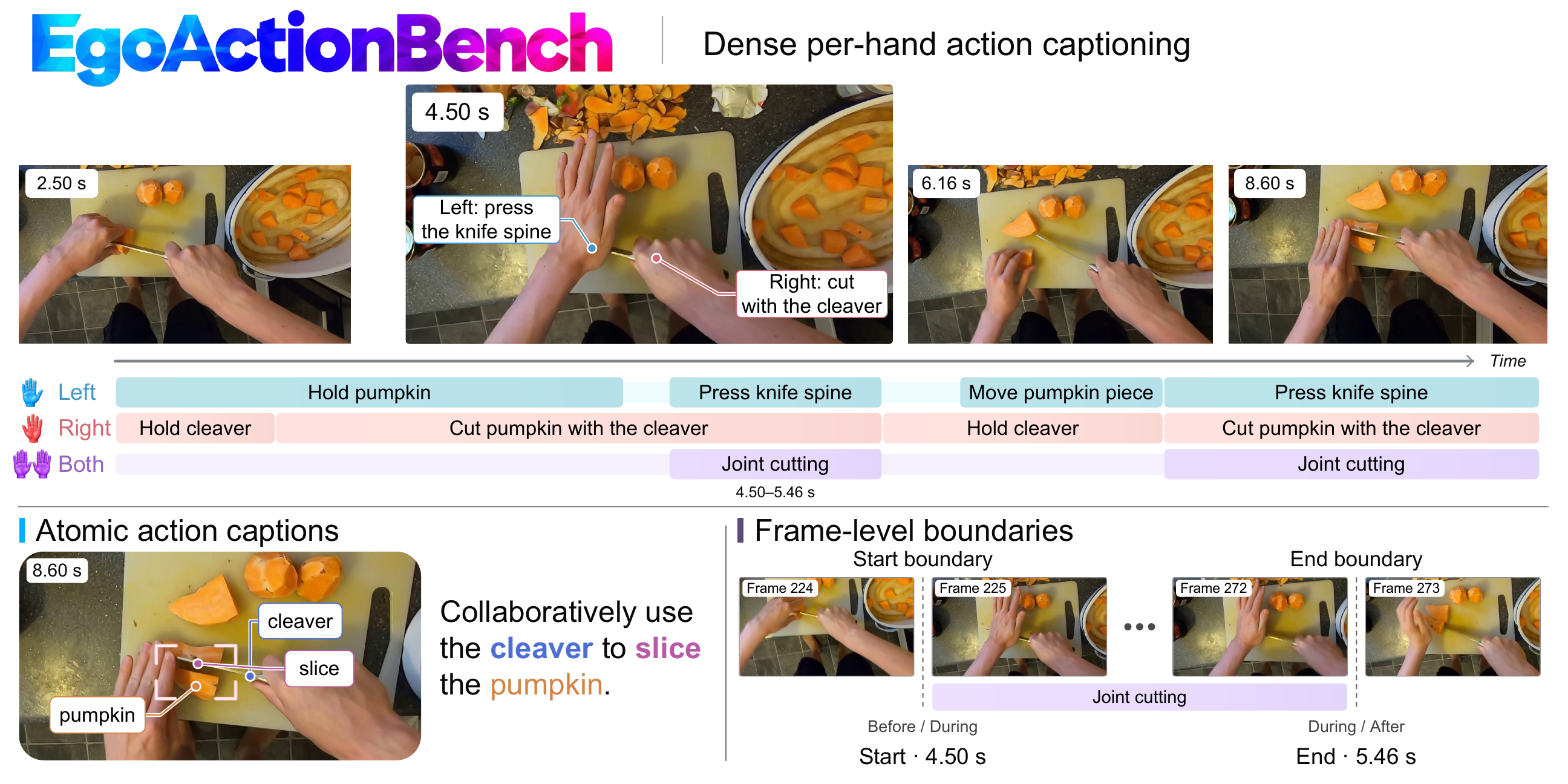}
\caption{\textbf{Overview of EgoAction.}
Continuous egocentric video is represented as temporally localized atomic
actions on left-hand, right-hand, and both-hands tracks (top).
The bottom-left example highlights the action verb, manipulated object, and tool.
Adjacent frames illustrate the start and end boundaries of a joint
action (bottom right).}
\label{fig:teaser}
\end{figure}

Learning from human data offers a scalable path toward embodied intelligence \citep{nair2023r3m,kareer2025egomimic,punamiya2026egoverse,cao2026acedata0,wang2026ego2robot,zheng2026egoscale,li2026openaoe,deng2026humannet,wang2026humanego,gao2026dreamdojo}. Egocentric video is particularly useful in this context, as it records human interactions with objects from the actor's viewpoint and provides rich observations of everyday manipulation \citep{damen2021rescaling,grauman2022ego4d,grauman2024egoexo4d,perrett2025hdepic,hoque2026egodex,yang2025egolife,li2026egolive,paul2026egooscar,feng2026eventkitchen}. However, extracting useful action supervision from such videos requires more than recognizing the overall activity. Models must recover the individual actions within continuous manipulation, determine when each action occurs, and identify which hand performs it.

Existing video understanding benchmarks capture only parts of this problem. Action recognition typically maps predefined video segments to action categories \citep{kay2017kinetics,goyal2017something,damen2021rescaling}, while video captioning often describes an entire clip or a relatively coarse event \citep{xu2016msrvtt,krishna2017dense,zhou2018youcook2,ohkawa2025exo2egodvc}. Temporal action localization requires models to recover action boundaries, but is commonly formulated over predefined action vocabularies or events at a coarser granularity \citep{cabaheilbron2015activitynet,zhang2022actionformer,wang2023egoonly}. Recent egocentric benchmarks have moved closer to understanding everyday human activities and hand--object interactions \citep{darkhalil2022visor,grauman2024egoexo4d,perrett2025hdepic,tateno2026handyvqa,leng2026flora}. However, these formulations do not jointly address three requirements for recovering fine-grained human manipulation: \textbf{action atomicity}, \textbf{temporal precision}, and \textbf{hand-action association}.

\textbf{Action atomicity} requires decomposing continuous manipulation into individual actions rather than summarizing a sequence with a single activity label. For example, an interval described as ``stirring'' may include picking up a spoon, stirring with it, and putting it down. Each action captures a distinct interaction with the object. A fine-grained description should preserve these actions and specify the objects involved, distinguishing, for example, stirring with a spoon from stirring with chopsticks.

\textbf{Temporal precision} requires identifying the start and end of each action, including actions that last less than a second. For example, a person washing dishes may briefly move the faucet handle to turn on the water. Missing this short action leaves a gap in the recovered sequence, even if the surrounding activity is correctly recognized as ``washing dishes.'' Recovering such actions requires sufficiently fine temporal resolution to distinguish them from adjacent actions and localize their boundaries.

\textbf{Hand-action association} requires assigning each action to the left hand, right hand, or both hands while preserving their temporal relationships. The two hands can perform different actions over different intervals: the left hand may hold a bowl throughout, while the right hand picks up a spoon, stirs, and puts it down. Describing the interval as ``holding a bowl and stirring'' obscures which hand performs each action and how the actions overlap. A fine-grained representation should retain the sustained left-hand action alongside the sequence of shorter right-hand actions.

To study this capability, we introduce \textsc{EgoActionBench}, a benchmark for fine-grained egocentric action understanding. \textsc{EgoActionBench} represents continuous human manipulation as temporally localized atomic actions associated with the left hand, right hand, or both hands. It contains 1,407 episodes with 41,132 human-refined action annotations. Rather than providing models with predefined temporal segments to classify or describe, the benchmark requires them to recover the action structure directly from continuous egocentric video. Its annotations capture short-lived actions and independent hand behaviors with fine temporal granularity, enabling systematic evaluation beyond coarse activity descriptions.

Evaluating such predictions also requires careful treatment of temporal segmentation. Predicted and reference actions can differ in both their temporal boundaries and semantic content. A single prediction may span multiple reference actions, while several predictions may fragment one reference action. Existing evaluation protocols based on merging or strict one-to-one matching for semantic scoring can therefore confound segmentation differences with semantic errors \citep{fujita2020soda,yao2026timechatcaptioner}. We propose a progressive evaluation protocol that first retrieves temporally eligible predictions, selects one by content agreement, and then scores its temporal extent and hand association. This ordering evaluates semantic fidelity over temporally corresponding actions while retaining the ability to diagnose localization errors. A separate one-to-one matching over positive-content candidate pairs measures action precision and recall, so that output density also contributes to the overall score.

We benchmark 20 baseline models and systems on \textsc{EgoActionBench}, spanning proprietary models, open-source general-purpose VLMs, video captioning models, and egocentric video understanding models, and analyze their ability to recover fine-grained actions from continuous egocentric video. Our experiments reveal substantial limitations in current models, including missed short actions, inaccurate temporal boundaries, merged action sequences, and incorrect associations between actions and hands. We further adapt two backbones, Qwen3-VL-4B and LLaVA-OneVision-2-8B, through supervised fine-tuning followed by reinforcement learning on automatically annotated training episodes, studying whether supervision at the target granularity can improve action understanding without modifying the underlying model architectures. Both adapted models outperform all evaluated open-source baselines. These results establish baselines for \textsc{EgoActionBench} and highlight the gap between general video understanding and the detailed action understanding needed to learn effectively from human data.

Our main contributions are:
\begin{itemize}
\item We introduce \textsc{EgoActionBench}, a benchmark containing 1,407 egocentric episodes and 41,132 human-refined action annotations, representing continuous human manipulation as temporally localized atomic actions associated with the left hand, right hand, or both hands.

\item We propose a progressive evaluation protocol that first retrieves temporally eligible predictions and then uses content agreement to select a shared correspondence for semantic fidelity, temporal overlap, and hand assignment. Separate action matching measures precision and recall, enabling fine-grained evaluation when predicted and reference actions differ in temporal segmentation.

\item We benchmark 20 baseline models and systems and adapt Qwen3-VL-4B and LLaVA-OneVision-2-8B through supervised fine-tuning followed by reinforcement learning using automatically annotated training data, revealing persistent challenges in recovering fine-grained action structure from continuous egocentric video.

\end{itemize}

\section{Related Work}
\label{sec:related_work}

\subsection{Egocentric Video Understanding}
Egocentric action understanding has progressed from transferable
video--language representations, to interaction-aware modeling, and finally to
language-grounded descriptions of ongoing activity. This progression is
relevant to embodied systems, where first-person observations
connect perception to the hand and object being manipulated. Large-scale
resources and narration-based pretraining established the foundation for
egocentric learning~\citep{damen2021rescaling,grauman2022ego4d,
grauman2024egoexo4d,lin2022egovlp,zhao2023lavila,pramanick2023egovlpv2}.
Subsequent work made hand--object relations, bimanual structure, motion, and
spatial cues more explicit through supervision and interaction-aware
models~\citep{darkhalil2022visor,zhang2022egohos,kwon2021h2o,
liu2024taco,pei2025egovideo,xu2025egodtm}. Video-language models
and cross-view methods use language for long-video description, question
answering, and knowledge transfer~\citep{ren2024timechat,ye2025mmego,
bai2025qwen3vl,zhang2026exo2ego,li2026egocross}.

\subsection{Video Captioning Benchmarks}

Video captioning benchmarks mainly differ in the granularity of the described
event, the output required from the model, and the evidence used for
evaluation. Early third-person benchmarks established dense captioning for
general events and procedural steps~\citep{krishna2017dense,zhou2018youcook2,
zhou2018endtoend}. Egocentric extensions vary between single-activity
captioning, hierarchical narration, and localized procedural description
\citep{nakamura2021sensoraugmented,islam2024videorecap,
ohkawa2025exo2egodvc,shi2025ue2dpac}. For embodied manipulation, WGO-Bench
adds instruction-conditioned timestamped subtask annotation, but does not
require an explicit acting-hand track~\citep{macrodata2026wgobench,
macrodata2026wgoreport}. Flora~\citep{leng2026flora} combines temporal
segmentation and contextual captioning to annotate intended actions in human
and robot videos. Its DenseCaptionBench evaluates temporal localization and
caption quality, including manipulator coordination. HanDyVQA is complementary: it evaluates
hand--object reasoning through question-conditioned QA rather than open-ended
caption generation and autonomous temporal discovery~\citep{tateno2026handyvqa}.

\begin{table}[!htbp]
\centering
\scriptsize
\caption{\textbf{Comparison with related datasets and benchmarks.}
Actions may group several operations under one label; atomic actions are delimited by meaningful changes in the operation, manipulated object, contact relation, or dominant motion.
Temporal localization and caption generation denote evaluated tasks.
Hand-specific captions identify acting hands; independent hand timelines assign separate action intervals to each hand.
Structured scoring evaluates semantic fields or facts.
Checks denote documented protocol features.
Rows use Ego4D narrations and Ego-Exo4D atomic descriptions.
Percentages denote human egocentric episodes.}
\label{tab:benchmark_positioning}
\setlength{\tabcolsep}{2pt}
\renewcommand{\arraystretch}{1.02}
\resizebox{\linewidth}{!}{%
\begin{tabular}{@{}l l l c c c c c c@{}}
\toprule
\textbf{Dataset / Benchmark} & \textbf{Scene} & \makecell{\textbf{Target}\\\textbf{unit}} & \textbf{Ego.} & \makecell{\textbf{Temp.}\\\textbf{loc.}} & \makecell{\textbf{Caption}\\\textbf{gen.}} & \makecell{\textbf{Hand-specific}\\\textbf{captions}} & \makecell{\textbf{Independent}\\\textbf{hand timelines}} & \makecell{\textbf{Structured}\\\textbf{caption scoring}} \\
\midrule
\mbox{ActivityNet Captions~\citep{krishna2017dense}} & Real-world & Events & \textcolor{red}{\ding{55}} & \textcolor{teal}{\ding{51}} & \textcolor{teal}{\ding{51}} & \textcolor{red}{\ding{55}} & \textcolor{red}{\ding{55}} & \textcolor{red}{\ding{55}} \\
\mbox{EPIC-KITCHENS-100~\citep{damen2021rescaling}} & Kitchen & Actions & \textcolor{teal}{\ding{51}} & \textcolor{teal}{\ding{51}} & \textcolor{red}{\ding{55}} & \textcolor{red}{\ding{55}} & \textcolor{red}{\ding{55}} & \textcolor{red}{\ding{55}} \\
\mbox{Ego4D~\citep{grauman2022ego4d}} & Real-world & Narrated events & \textcolor{teal}{\ding{51}} & \textcolor{red}{\ding{55}} & \textcolor{red}{\ding{55}} & \textcolor{red}{\ding{55}} & \textcolor{red}{\ding{55}} & \textcolor{red}{\ding{55}} \\
\mbox{Ego-Exo4D~\citep{grauman2024egoexo4d}} & Real-world & \makecell[l]{Atomic\\descriptions} & \textcolor{teal}{\ding{51}} & \textcolor{red}{\ding{55}} & \textcolor{red}{\ding{55}} & \textcolor{teal}{\ding{51}} & \textcolor{red}{\ding{55}} & \textcolor{red}{\ding{55}} \\
\mbox{EgoYC2~\citep{ohkawa2025exo2egodvc}} & Kitchen & Procedural steps & \textcolor{teal}{\ding{51}} & \textcolor{teal}{\ding{51}} & \textcolor{teal}{\ding{51}} & \textcolor{red}{\ding{55}} & \textcolor{red}{\ding{55}} & \textcolor{red}{\ding{55}} \\
\mbox{HD-EPIC~\citep{perrett2025hdepic}} & Kitchen & Actions & \textcolor{teal}{\ding{51}} & \textcolor{red}{\ding{55}} & \textcolor{red}{\ding{55}} & \textcolor{teal}{\ding{51}} & \textcolor{red}{\ding{55}} & \textcolor{red}{\ding{55}} \\
\mbox{VidCapBench~\citep{chen2025vidcapbench}} & Real-world & Clips & \textcolor{red}{\ding{55}} & \textcolor{red}{\ding{55}} & \textcolor{teal}{\ding{51}} & \textcolor{red}{\ding{55}} & \textcolor{red}{\ding{55}} & \textcolor{teal}{\ding{51}} \\
\mbox{HanDyVQA~\citep{tateno2026handyvqa}} & Real-world & HOI questions & \textcolor{teal}{\ding{51}} & \textcolor{red}{\ding{55}} & \textcolor{red}{\ding{55}} & \textcolor{red}{\ding{55}} & \textcolor{red}{\ding{55}} & \textcolor{red}{\ding{55}} \\
\mbox{WGO-Bench~\citep{macrodata2026wgobench}} & Manipulation & \makecell[l]{Manipulation\\events} & 25\% & \textcolor{teal}{\ding{51}} & \textcolor{teal}{\ding{51}} & \textcolor{red}{\ding{55}} & \textcolor{red}{\ding{55}} & \textcolor{red}{\ding{55}} \\
\mbox{OmniDCBench~\citep{yao2026timechatcaptioner}} & Real-world & Scenes & \textcolor{red}{\ding{55}} & \textcolor{teal}{\ding{51}} & \textcolor{teal}{\ding{51}} & \textcolor{red}{\ding{55}} & \textcolor{red}{\ding{55}} & \textcolor{teal}{\ding{51}} \\
\mbox{CAPEval~\citep{liu2026capeval}} & Diverse & Images & \textcolor{red}{\ding{55}} & \textcolor{red}{\ding{55}} & \textcolor{teal}{\ding{51}} & \textcolor{red}{\ding{55}} & \textcolor{red}{\ding{55}} & \textcolor{teal}{\ding{51}} \\
\mbox{EgoVerse-I~\citep{punamiya2026egoverse}} & Manipulation & \makecell[l]{Fine-grained\\segments} & \textcolor{teal}{\ding{51}} & \textcolor{red}{\ding{55}} & \textcolor{red}{\ding{55}} & \textcolor{teal}{\ding{51}} & \textcolor{red}{\ding{55}} & \textcolor{red}{\ding{55}} \\
\mbox{DenseCaptionBench~\citep{leng2026flora}} & Manipulation & Actions & 33.3\% & \textcolor{teal}{\ding{51}} & \textcolor{teal}{\ding{51}} & \textcolor{teal}{\ding{51}} & \textcolor{red}{\ding{55}} & \textcolor{teal}{\ding{51}} \\
\midrule
\mbox{\textbf{EgoActionBench (ours)}} & \textbf{Real-world} & \textbf{Atomic actions} & \textcolor{teal}{\ding{51}} & \textcolor{teal}{\ding{51}} & \textcolor{teal}{\ding{51}} & \textcolor{teal}{\ding{51}} & \textcolor{teal}{\ding{51}} & \textcolor{teal}{\ding{51}} \\
\bottomrule
\end{tabular}%
}
\end{table}

On the other hand, evaluation protocols range from caption similarity and temporal localization
to factuality checks and question-answer accuracy
\citep{liu2023factvc,chen2025vidcapbench,yao2026timechatcaptioner}.
EgoActionBench takes a full egocentric clip without action queries or
reference boundaries. It requires atomic action sequences on explicit Left,
Right, and Both tracks, with temporal alignment evaluated against frame-level
reference boundaries and hand assignment scored separately.

\section{EgoActionBench}
\label{sec:benchmark}

EgoActionBench evaluates dense hand-action captioning in egocentric videos.
We first define the task and its output format, then describe data collection
and annotation, and finally present a GT-anchored evaluation protocol that
separately measures semantic content, temporal localization, and hand assignment.

\subsection{Task Definition}
\label{sec:task}

We introduce the task of \emph{dense per-hand action captioning}
for egocentric video understanding.
Given a complete egocentric video clip, a model must temporally
localize and describe the camera wearer's visible atomic hand
actions, with an explicit hand assignment for each action.

\paragraph{Action vs.\ Atomic Action.}
We distinguish actions from atomic actions by their semantic granularity.
An action represents a goal-directed manipulation event that may involve
multiple operational phases. In contrast, an atomic action represents the
smallest temporally continuous and semantically coherent unit of observable
manipulation, bounded by meaningful changes in the operation, target object,
contact relation, or dominant motion. For example, \emph{picking up a cup}
constitutes an action, whereas \emph{reaching for the cup}, \emph{grasping it},
and \emph{lifting it} are distinct atomic actions. Thus, actions are organized
around functional goals, whereas atomic actions are grounded in observable
interaction dynamics.

Each prediction is represented as
\begin{equation}
p_i=(\hat{s}_i,\hat{e}_i,\hat{h}_i,w_i),
\label{eq:prediction}
\end{equation}
where $[\hat{s}_i,\hat{e}_i)$ is the predicted temporal interval,
$\hat{h}_i\in\{L,R,B\}$ denotes the Left, Right, or Both track,
and $w_i$ is a free-text action caption.
No action queries or reference boundaries are provided.

Each caption describes one atomic action.
The Left and Right tracks describe actions of the corresponding
anatomical hands.
The Both track captures coordinated manipulation of a shared
object, rather than simultaneous independent actions.
Both captions describe the collaboration, while any annotated
single-hand roles remain separate records on Left and Right.

\subsection{Data Collection}
\label{sec:benchmark_data}

\paragraph{Data sources and selection.}
Episodes are curated from Ego4D~\citep{grauman2022ego4d},
EgoLife~\citep{yang2025egolife}, and
EPIC-KITCHENS-100~\citep{damen2021rescaling}.
We combine these sources to cover hand-object interactions across kitchen,
household, and occupational settings.
Each episode captures a coherent task or subtask performed by the
camera wearer and is screened for suitability for dense hand-object
annotation. Sufficient surrounding context is retained to distinguish the target operations
from neighboring activities. Episodes with severe occlusion, unverifiable
actions, or unrelated task transitions are discarded.
Source-video identifiers and original temporal ranges are retained to trace
each episode and support separation from the adaptation data.
Additional dataset details and annotated examples are provided in
\cref{app:data_collection,app:dataset_examples}.

\paragraph{Dataset statistics.}
\label{sec:benchmark_statistics}
The collection contains 1,407 episodes with 41,132 action captions across the
Left, Right, and Both tracks (\cref{tab:benchmark_statistics}).
It spans 7.982 hours, with a mean episode duration of 20.42 seconds and a median
action duration of 0.60 seconds. Non-action entries, including invisibility,
idle intervals, and the absence of bimanual collaboration, are excluded from
the action counts.

\begin{table}[t]
\centering
\footnotesize
\setlength{\tabcolsep}{3.7pt}
\renewcommand{\arraystretch}{0.90}
\setlength{\abovecaptionskip}{4pt}
\caption{\textbf{Dataset Statistics of EgoActionBench.} Video clips, total video duration, and human-refined action annotations by source. Left, Right, and Both report annotation counts for the corresponding hand tracks.}
\label{tab:benchmark_statistics}
\scalebox{0.94}{\begin{tabular}{lrrrrrr}
\toprule
Source & Clips & Hours & Actions & Left & Right & Both \\
\midrule
Ego4D & 451 & 3.130 & 16,774 & 6,825 & 8,349 & 1,600 \\
EgoLife & 485 & 2.676 & 13,306 & 5,895 & 6,109 & 1,302 \\
EPIC-KITCHENS-100 & 471 & 2.176 & 11,052 & 4,599 & 5,300 & 1,153 \\
\midrule
Total & 1,407 & 7.982 & 41,132 & 17,319 & 19,758 & 4,055 \\
\bottomrule
\end{tabular}}
\end{table}

\begin{figure}[htbp]
\centering
\includegraphics[width=0.8\linewidth]{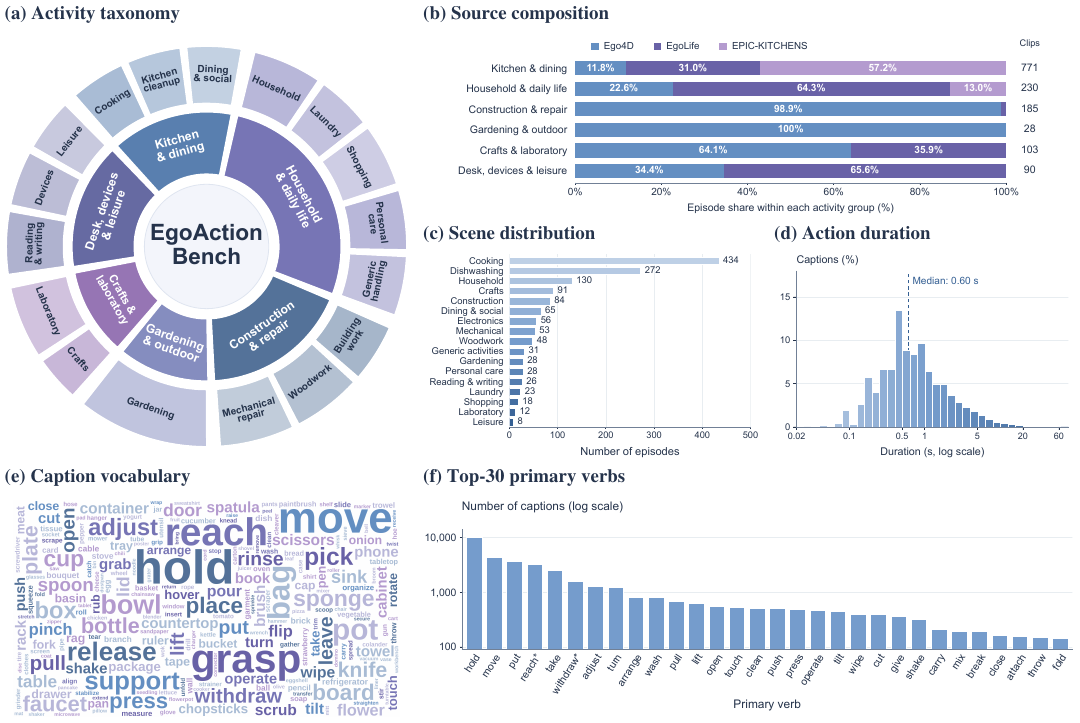}
\caption{\textbf{Dataset Composition and Action Statistics.}
(a) Activity taxonomy with six groups and 17 scenes.
(b) Source composition within each activity group. (c) Episode counts by scene.
(d) Action-duration distribution on a logarithmic time axis, with a median
of 0.60\,s. (e) Caption vocabulary. (f) The 30 most frequent primary verbs,
with counts on a logarithmic scale. Statistics cover 1,407 episodes and
41,132 action captions; sector angles do not encode counts.}
\label{fig:data_statistics}
\end{figure}

\subsection{Data Annotation Pipeline}
\label{sec:benchmark_annotation}
\begin{figure}[htbp]
\centering
\includegraphics[width=0.8\linewidth]{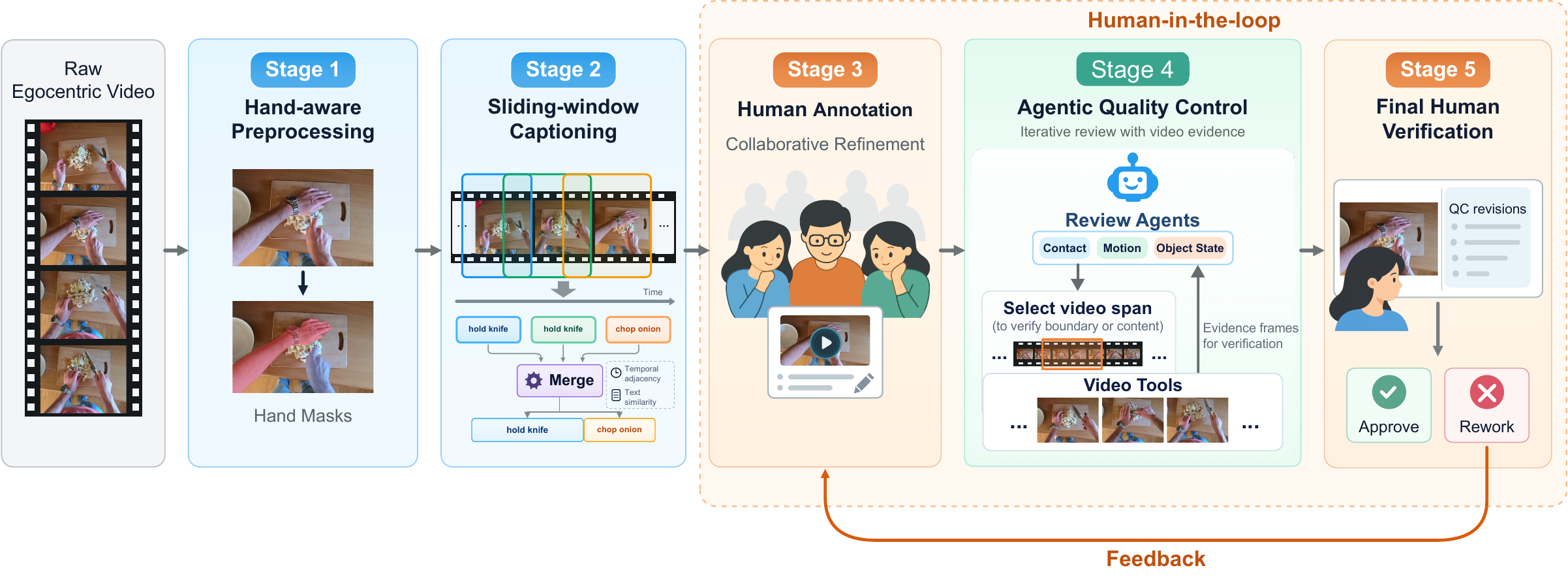}
\caption{\textbf{Human-in-the-Loop Construction of EgoActionBench.}
Hand-aware preprocessing and sliding-window captioning generate draft
per-hand annotations. Human annotation, agentic quality control, and
final human verification refine these drafts against the original video.
Episodes needing correction return for human annotation and QC.}
\label{fig:annotation_pipeline}
\label{fig:data_pipeline}
\end{figure}
We annotate the selected episodes through a five-stage human-in-the-loop
pipeline (\cref{fig:annotation_pipeline}).
To help the captioning model associate each action with the correct
hand, we extract left and right hand masks using
CaRe-Ego~\citep{su2025careego}. The masks are temporally smoothed and
rendered as distinct overlays on the video.
A VLM then generates sliding-window captions for the Left, Right, and Both
tracks, merging consecutive fragments of the same continuous action into
draft annotations. Human annotators correct action descriptions, hand
assignments, and temporal boundaries against the original video.
Agentic quality control subsequently inspects local visual evidence to
propose further revisions, guided by hand-object contact, hand-motion phases,
and object-state changes. Finally, human reviewers verify the annotations;
episodes requiring further correction return for annotation and QC.
Only human-verified annotations enter the benchmark.
Detailed annotation and quality-control procedures are as follows.

\paragraph{Stage 1: Hand-aware preprocessing.}
We extract left and right hand masks using
CaRe-Ego~\citep{su2025careego}. Temporal smoothing filters brief or inconsistent
detections using the continuity of hand regions across neighboring frames.
The filtered masks are rendered as distinct overlays that highlight each
hand while preserving the surrounding objects and scene. This hand-highlighted
video provides explicit visual cues for the subsequent per-hand captioning.

\paragraph{Stage 2: Sliding-window captioning.}
A VLM processes the hand-highlighted video in sliding windows and produces
one-second caption proposals for the Left, Right, and Both tracks. It uses
frames from the current window and recent captions as context. To join
actions that span multiple windows, we identify candidate merges within
each track using temporal adjacency and text similarity. Candidate merges
are then checked against the video to confirm that the fragments describe
the same uninterrupted action. Segments separated by task-relevant changes in the operation,
manipulated object, or contact relation are kept distinct. The resulting
variable-length segments serve as drafts for human annotation.

\paragraph{Stage 3: Human annotation.}
Annotators inspect the original video and revise the drafts according to
the task definition in \cref{sec:task}. They correct action descriptions and
hand assignments, add omitted actions, split compound events, and remove
details not supported by the video. Adjacent fragments are merged when they
describe the same uninterrupted action. The three tracks are reviewed
separately, since one hand may sustain an action while the other performs
several shorter operations. Annotators inspect frames from the original
video around action transitions to refine start and end times.

\paragraph{Stage 4: Agentic quality control.}
Reviewing dense annotations requires checking many action segments and
closely inspecting brief transitions. We use a team of agents to review
these annotations, with each agent checking assigned intervals for errors
in action descriptions, hand assignments, and temporal boundaries. Agents
inspect timestamped frames from the original video. Around ambiguous
boundaries, they examine consecutive native frames one by one to determine
when an action starts or ends. Three visual cues guide this review:
\begin{itemize}
    \item \textbf{Hand-object contact:} changes in contact or control.
    \item \textbf{Hand-motion phase:} changes between operations,
    even while maintaining a stable grasp.
    \item \textbf{Object-state change:} visible responses such as displacement,
    deformation, or content transfer.
\end{itemize}
Caption revisions correct the described operation, participating objects,
or hand roles. Boundary adjustments align action starts and ends with the
first visible evidence of the corresponding transition. Each proposed
revision includes supporting frame references and a brief explanation of
the change, allowing human reviewers to verify it against the original
video in Stage 5. Unresolved cases are flagged for further human review.

\paragraph{Stage 5: Final human verification.}
Human reviewers check the annotations against the original video, using
the proposed revisions, supporting frames, and explanations from Stage 4.
They resolve remaining ambiguities and either approve the annotations or
return the episode to Stage 3 for correction and another round of agentic
QC. Only approved annotations enter the benchmark.

\subsection{Evaluation Suite}
\label{sec:benchmark_metrics}

\begin{figure}[htbp]
\centering
\includegraphics[width=\linewidth]{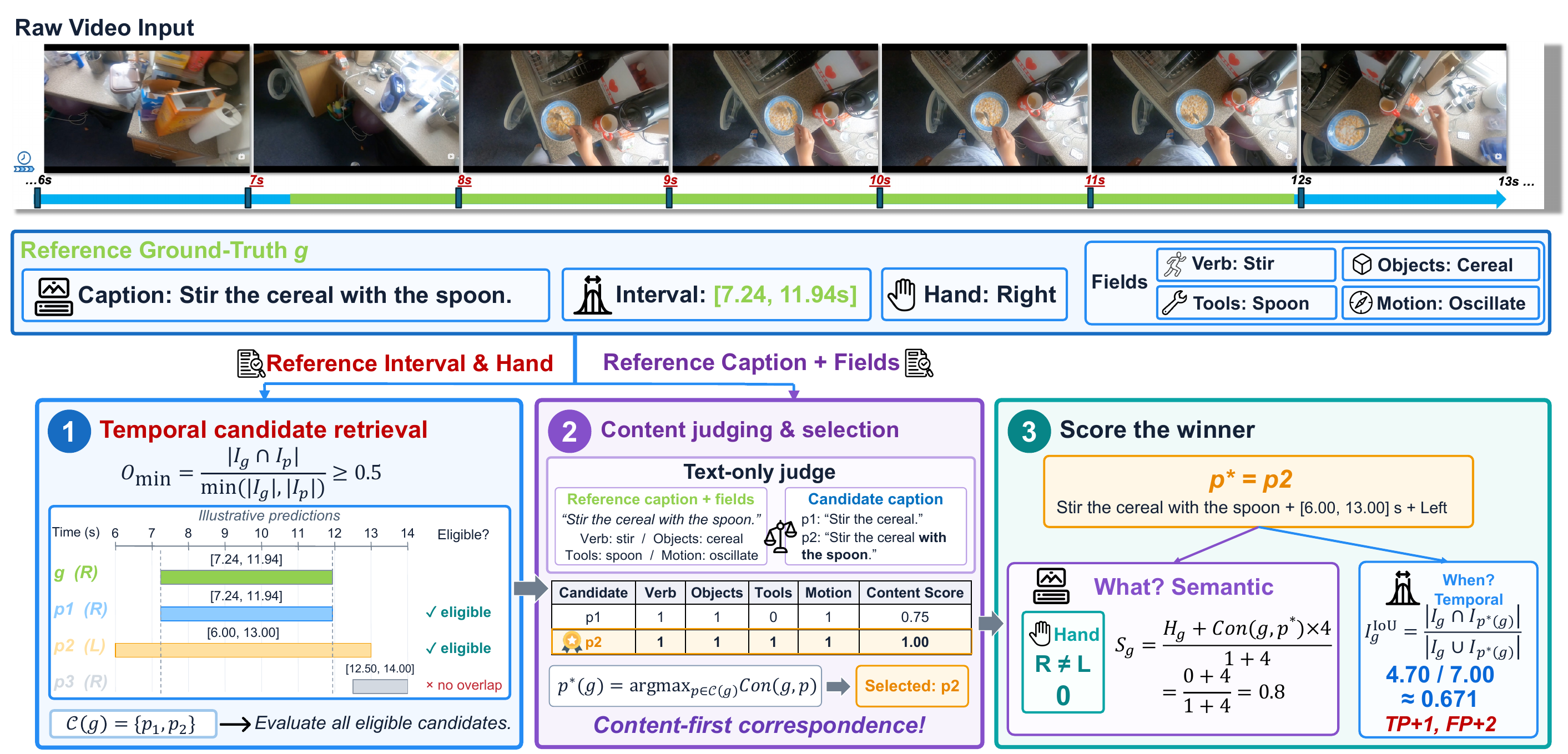}
\caption{\textbf{EgoAction evaluation pipeline.}
For each ground-truth action, temporal overlap and hand-track compatibility
determine the eligible candidates. Positive field-level semantic agreement selects
one prediction, whose original caption, interval, and hand label are used
to compute Semantic and temporal IoU. These scores are averaged separately
over ground-truth actions with completed evaluations.}
\label{fig:evaluation_suite}
\end{figure}

We evaluate \textbf{what} action is described, \textbf{when} it occurs, and
\textbf{which hand} performs it through Semantic and Temporal scores.
Our evaluation is explicitly \textbf{ground-truth-based}: each ground-truth
action selects at most one prediction, which supplies its Semantic score
and temporal IoU. As illustrated in \cref{fig:evaluation_suite}, we first retrieve
temporally eligible predictions from compatible hand tracks, then select
the candidate with the highest positive semantic agreement. A language-model judge
assesses semantic agreement across four fields: Verb, Objects, Tools, and
Motion. The selected prediction's semantic assessment and hand-label
agreement are combined into Semantic, while its interval determines
temporal IoU, with its caption and interval retained in full.
Semantic and temporal IoU are averaged
over ground-truth actions with completed evaluations on a $0$--$100$ scale.
Temporal averages the resulting m-tIoU and the micro-averaged F1 score
based on one-to-one action matching.
This combination captures interval overlap quality, action segmentation
completeness, and prediction redundancy.
The arithmetic mean of Semantic and Temporal defines
\textbf{EgoActionScore}. Full matching rules, scoring definitions, and
metric interpretation are provided as follows.

\paragraph{Evaluation inputs.}
Each ground-truth action and prediction are represented as
\begin{equation}
g=(I_g,h_g,c_g,F_g),
\qquad
p=(I_p,\hat{h}_p,\hat{c}_p),
\label{eq:gt_event}
\end{equation}
where $I$ is the temporal interval, $h\in\{L,R,B\}$ is the acting hand,
and $c$ is the natural-language caption. The structured description $F_g$
is provided only on the ground-truth side for semantic evaluation.
Models submit an interval, a hand track, and a caption; they are not
required to predict structured fields.

\paragraph{Stage 1: Candidate retrieval.}
For each ground-truth action $g$, candidates are drawn from predictions
in the same video. The admissible hand tracks are
\begin{equation}
\mathcal{H}(h_g)=
\begin{cases}
\{L,R\}, & h_g\in\{L,R\},\\
\{B\}, & h_g=B.
\end{cases}
\end{equation}
A single-hand action can retrieve predictions from either single-hand
track, keeping otherwise correct predictions eligible despite
left--right attribution errors. Bimanual and single-hand actions are
kept separate because the Both track represents joint two-hand actions.

Temporal eligibility is determined by overlap normalized by the shorter
interval:
\begin{equation}
O_{\min}(g,p)=
\frac{|I_g\cap I_p|}{\min(|I_g|,|I_p|)}.
\label{eq:candidate_overlap}
\end{equation}
Let $\mathcal{P}$ denote the predictions in the current video. The eligible
candidate set $\mathcal{C}(g)$ is
\begin{equation}
\mathcal{C}(g)=
\left\{
p\in\mathcal{P}
\;\middle|\;
\hat{h}_p\in\mathcal{H}(h_g),\;
O_{\min}(g,p)\geq\tau_{\mathrm{sem}}
\right\}.
\end{equation}
The default setting uses $\tau_{\mathrm{sem}}=0.5$ and evaluates
every eligible candidate; threshold sensitivity is considered in
\cref{sec:ablation_studies}. This criterion admits both a long prediction
containing a short ground-truth interval and a short prediction contained
within a long ground-truth interval. This retrieval criterion keeps predictions
with imperfect temporal boundaries eligible for semantic evaluation.
Temporal IoU separately measures interval alignment, assigning lower scores
to predictions that extend beyond or cover only part of the ground-truth interval.

\paragraph{Stage 2: Content judging.}
We assess action content through the operation, participating objects,
tools, and motion. Hand assignment is evaluated by directly comparing the
submitted track with the ground-truth hand label and is incorporated into
Semantic in Stage 3.
The structured reference $F_g=(v_g,O_g,U_g,\mu_g)$ contains four fields:
\textbf{Verb}, the core operation; \textbf{Objects}, the participating
entities and their action-dependent roles; \textbf{Tools}, the intermediaries
used to perform the operation; and \textbf{Motion}, the coarse physical
movement underlying the operation. Objects and Tools are distinguished
by their role: a knife is a tool when used for cutting, but an object when
it is picked up. Motion labels are
\texttt{reach}, \texttt{translate}, \texttt{lift}, \texttt{lower},
\texttt{rotate}, \texttt{press}, \texttt{oscillate}, or \texttt{static}.

For every $p\in \mathcal{C}(g)$, a language-model judge receives the ground-truth
caption $c_g$, structured fields $F_g$, and full predicted caption
$\hat{c}_p$, with no access to video frames. Required content is assessed
using direct quotes from the prediction. Equivalent wording receives
credit, while missing or contradictory content does not. Supporting
evidence must describe the same action and cannot combine distinct actions
within a caption.

Verb and Motion receive binary scores. Objects and Tools are evaluated
item by item and averaged within each field. Only fields marked as required
in the ground truth contribute; skipped and inapplicable fields are excluded.
Required content omitted
from a prediction receives zero credit and remains in the denominator.

Let $K_g$ denote the scored fields and $r_k(g,p)\in[0,1]$ their individual
scores. These four fields capture the semantic content of an action.
We average their applicable scores to obtain the \textbf{content score $Con(g,p)$}  used for
prediction selection:
\begin{equation}
Con(g,p)=\frac{1}{|K_g|}\sum_{k\in K_g}r_k(g,p).
\label{eq:semantic_pair}
\end{equation}
A field therefore receives no additional weight for containing more
items. For every eligible pair, the judge must resolve each required
field and its action binding.

\paragraph{Selecting one prediction per ground-truth action.}
For each candidate set containing a prediction with a positive content score,
we select the prediction with the highest content score:
\begin{equation}
p^*(g)=\operatorname*{arg\,max}_{p\in \mathcal{C}(g),\,Con(g,p)>0}Con(g,p).
\label{eq:selection}
\end{equation}
Among candidates with equal content scores, we prioritize agreement with
the ground-truth hand, followed by higher temporal IoU.
Since hand identity is part of the action's semantics, this order favors
semantic fidelity before temporal alignment.
Each ground-truth action selects at most one prediction,
but the same prediction may be selected for multiple ground-truth actions.
We retain the selected caption and interval without combining evidence
across predictions or cropping temporal boundaries.

\paragraph{Stage 3: Component scoring.}
Using the selected prediction, we compute hand-label agreement, Semantic,
and temporal IoU:
\begin{equation}
H_g=\mathbf{1}[h_g=\hat{h}_{p^*(g)}],
\qquad
S_g=\frac{H_g+\sum_{k\in K_g}r_k(g,p^*(g))}{1+|K_g|},
\end{equation}
\begin{equation}
I_g^{\mathrm{IoU}}=
\frac{|I_g\cap I_{p^*(g)}|}{|I_g\cup I_{p^*(g)}|}.
\label{eq:tiou}
\end{equation}
Temporal IoU uses the same selected prediction
and its original interval. Actions without a candidate with a positive content score
receive zero for all three quantities and remain in the averaging denominator.

\begin{figure}[htbp]
\centering
\includegraphics[width=\linewidth]{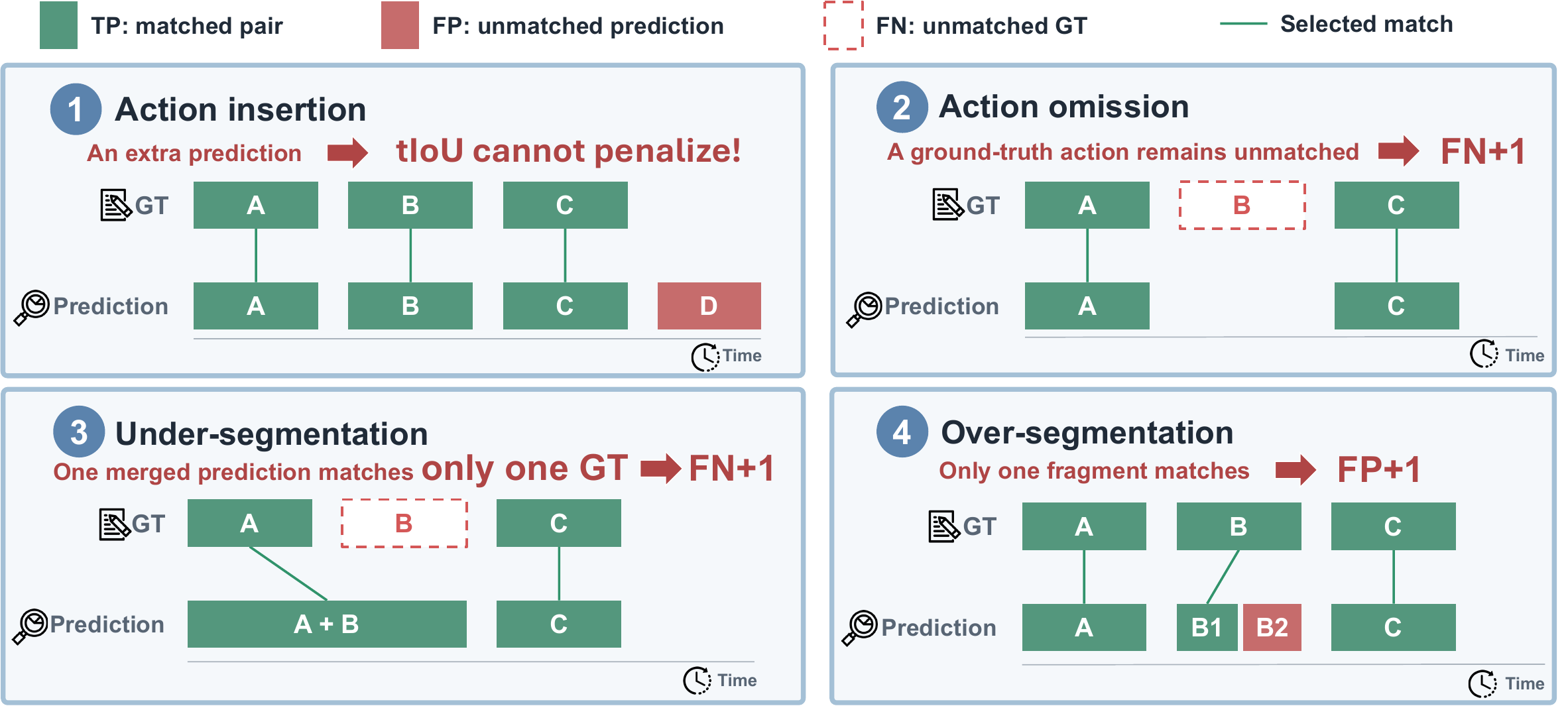}
\caption{\textbf{Why F1 score matters.}
We complement tIoU, which measures interval overlap, with one-to-one F1
  to explicitly penalize errors in event coverage and segmentation
  granularity. The four schematic examples illustrate
  (1) action insertion, (2) action omission,
  (3) under-segmentation, and
  (4) over-segmentation.}
\label{fig:why_f1_matters}
\end{figure}

\textbf{Action matching.}
Temporal IoU measures interval overlap, but does not by itself enforce a
one-to-one correspondence or explicitly penalize every unmatched prediction.
We therefore complement tIoU with an action-level F1 score to evaluate event
coverage and segmentation granularity (Figure~\ref{fig:why_f1_matters}).
For each video $v$, we find a maximum-cardinality one-to-one matching $M_v$
over all eligible pairs $(g,p)$ with $Con(g,p)>0$. Each matched pair
contributes one true positive. We pool counts across videos to compute
the micro-averaged F1 score:
\begin{equation}
TP=\sum_v|M_v|,\qquad FP=N_P-TP,\qquad FN=N_G-TP,
\end{equation}
\begin{equation}
\mathrm{F1}=100 \times \frac{2TP}{N_P+N_G}.
\label{eq:action_f1}
\end{equation}
Here, $N_G$ is the number of evaluated ground-truth actions, and $N_P$
counts all identifiable submitted predictions in the input videos,
including unmatched predictions and records with invalid intervals. One-to-one action matching
penalizes over-segmentation through false positives and under-segmentation
through false negatives.

Let $\mathcal{G}$ denote all ground-truth actions and
$\mathcal{G}_{\mathrm{eval}}\subseteq\mathcal{G}$ the actions with all candidate judgments
completed for a given model. Actions without a candidate with a positive content score receive
zero scores and are included in $\mathcal{G}_{\mathrm{eval}}$.
All components use this same evaluated ground-truth set, with
$N_G=|\mathcal{G}_{\mathrm{eval}}|$. Each evaluated action receives equal
weight, with scores on a $0$--$100$ scale:
\begin{equation}
(\mathrm{Semantic},\mathrm{m\mbox{-}tIoU},\mathrm{Hand})
=
\frac{100}{N_G}
\sum_{g\in\mathcal{G}_{\mathrm{eval}}}(S_g,I_g^{\mathrm{IoU}},H_g).
\end{equation}
Temporal averages m-tIoU and F1, and the arithmetic mean of Semantic and
Temporal defines the aggregate score:
\begin{equation}
\mathrm{Temporal}=\frac{\mathrm{m\mbox{-}tIoU}+\mathrm{F1}}{2},
\end{equation}
\begin{equation}
\mathrm{EgoActionScore}
=
\frac{\mathrm{Semantic}+\mathrm{Temporal}}{2}.
\label{eq:egoaction_score}
\end{equation}
We additionally report Hand and the four content-field scores, with each
content field averaged over ground-truth actions requiring that field.

\paragraph{Metric interpretation.}
EgoActionScore combines semantic fidelity, temporal overlap, and action
precision and recall. A prediction may support multiple ground-truth
actions in semantic scoring but can match at most one action when
computing F1. Additional unmatched predictions reduce F1.
Hand scores should also be interpreted
in light of candidate filtering. Since a Both ground-truth action admits
only Both predictions, its Hand score reflects whether a candidate with a
positive content score is available, rather than unrestricted three-way hand
classification.

\section{Learning to Caption Egocentric Actions}
\label{sec:models}

We study whether supervision at the granularity defined in
\cref{sec:benchmark} improves fine-grained egocentric action understanding.
Given a video, the model must recover individual actions, their temporal
boundaries, and the hands that perform them. EgoAction learns this mapping through
supervised fine-tuning followed by reinforcement learning.

\paragraph{Training data.}
\label{sec:sft_data}
The training set contains 5,354 egocentric video episodes drawn from
Ego4D~\citep{grauman2022ego4d},
EPIC-KITCHENS~\citep{damen2021rescaling}, and
EgoLife~\citep{yang2025egolife}, including 5,146 training examples and
208 validation examples.
Each example pairs a video with timestamped captions organized into
Left, Right, and Both tracks. The targets jointly supervise action descriptions,
temporal intervals, and hand assignments. Unlike the human-refined annotations
used to evaluate EgoActionBench, these training labels are generated and refined
automatically. Source identifiers are retained to separate the adaptation data
from the benchmark evaluation set.

\paragraph{Supervised adaptation.}
We perform supervised fine-tuning on two pretrained backbones:
Qwen3-VL-4B-Instruct~\citep{bai2025qwen3vl} and
LLaVA-OneVision-2-8B~\citep{an2026llavaov2}.
Qwen3-VL serves as a representative general-purpose model for
image and video understanding. LLaVA-OneVision-2 provides a
complementary backbone through its native-resolution
OneVision-Encoder, unified spatiotemporal modeling, and emphasis
on temporal and spatial grounding.
This pairing allows us to examine whether dense action
supervision benefits different pretrained models, including one
already oriented toward fine-grained video perception.
Given a video and a task instruction, each model is trained to
generate timestamped action captions across the Left, Right,
and Both tracks. Action descriptions, temporal intervals, and
hand assignments are jointly supervised through next-token
prediction, with the loss applied only to the target response.
The original architectures are retained, with the visual
encoders frozen and the language models and multimodal
connectors updated.

\paragraph{Reinforcement learning.}
To optimize semantic fidelity, temporal alignment, and action coverage at the
sequence level, we further refine the supervised models using group relative
policy optimization (GRPO; \citealp{shao2024deepseekmath}). For each training
video $v$, the model generates $G$ alternative sets of timestamped action
captions, denoted by $\{y_i\}_{i=1}^{G}$. GRPO uses each output's reward
relative to the group mean to guide policy updates, favoring higher-reward
predictions. A KL penalty against the SFT policy regularizes these updates.

Each candidate output is scored against the training reference using the
following reward:
\begin{equation}
R(v,y_i)
=0.5\,S(v,y_i)+0.25\,I(v,y_i)+0.25\,F(v,y_i),
\label{eq:grpo_reward}
\end{equation}
where $S$ measures semantic fidelity, including hand assignment, $I$ measures
temporal overlap, and $F$ is F1 from one-to-one action matching, accounting
for missed and redundant predictions.

\raggedbottom
\section{Experiments}
\label{sec:experiments}
\subsection{Experimental Setup}
\label{sec:experimental_setup}

We evaluate 20 baseline models and systems on EgoActionBench, spanning four categories: proprietary models, open-source general-purpose VLMs, video captioning models, and egocentric video understanding models. This comparison examines how general multimodal
capability and task specialization translate to dense action captioning.
The EgoAction models introduced in Section~\ref{sec:models} are evaluated alongside their
Qwen3-VL-4B and LLaVA-OneVision-2-8B backbones.
Models receive video inputs through their respective input interfaces and
are prompted to follow the task specification in Section~\ref{sec:task}.
Evaluation follows the suite in Section~\ref{sec:benchmark_metrics}.
The open-weight Qwen3-32B~\citep{yang2025qwen3} serves as the semantic judge,
supporting reproducible evaluation with a fixed model checkpoint.
Reported metrics include EgoActionScore, its Temporal and Semantic
components, and detailed scores for Hand, Verb, Objects, Tools, and Motion.

\paragraph{Human baseline.}
To establish a human baseline, 3 annotators independently annotate
evaluation episodes following the task definition in Section~\ref{sec:task}.
They identify action boundaries, assign hand tracks, and write action
captions from the original videos, without access to the reference
annotations or model predictions.
Their annotations are evaluated against the fixed benchmark references
using the same matching procedure, semantic judge, and scoring rules
as the models.
Table~\ref{tab:main_results} reports their scores as Human Expert.

\subsection{Main Benchmark Results}
\label{sec:main_results}

\paragraph{Dense action captioning remains challenging for current VLMs.}
Table~\ref{tab:main_results} reveals a substantial gap between current models
and human performance: even GPT-6 Astra~\citep{openai2026gpt6astra}, the strongest model evaluated,
achieves an EgoActionScore of 59.6 compared with 90.4 for human experts.
Performance does not increase monotonically with model size, and models
specialized for video captioning or egocentric understanding do not
consistently outperform general-purpose VLMs. These results highlight the
difficulty of jointly recovering action content, temporal boundaries,
and hand assignments.

\begin{table}[!htbp]
\centering
\scriptsize
\setlength{\tabcolsep}{4pt}
\renewcommand{\arraystretch}{1.0}
\newcommand{\modelgroup}[1]{\multicolumn{11}{@{}l}{\textit{#1}}}
\caption{\textbf{Dense action captioning on EgoActionBench.}
All scores range from 0 to 100; higher is better.
Temporal Overall is the mean of m-tIoU and micro-averaged F1.
Semantic Overall averages the applicable Hand, Verb, Objects, Tools, and Motion scores within each ground-truth action, then averages over ground-truth actions.
EgoActionScore is the mean of Temporal Overall and Semantic Overall.
Parentheses identify the EgoAction backbone.
Flora uses a 9B temporal segmenter followed by a 27B captioner.
Best scores in each column are bolded separately for proprietary and open-source models.}
\label{tab:main_results}
\begin{adjustbox}{max width=\linewidth}
\begin{tabular}{@{}l r @{\hspace{7pt}} rrr @{\hspace{7pt}} rrrrrr@{}}
\toprule
\multirow{2}{*}{\textbf{Model / system}}
& \multirow{2}{*}{\makecell{\textbf{EgoAction}\\\textbf{Score}}}
& \multicolumn{3}{c}{Temporal} & \multicolumn{6}{c}{Semantic} \\
\cmidrule(lr){3-5}\cmidrule(lr){6-11}
& & m-tIoU & F1 & Overall
& Hand & Verb & Objects & Tools & Motion & Overall \\
\midrule
\modelgroup{Human Baseline} \\
\midrule
Human Expert & 90.4 & 88.4 & 93.1 & 90.8 & 95.5 & 94.5 & 95.1 & 94.8 & 74.5 & 90.1 \\
\midrule
\modelgroup{Proprietary models} \\
\addlinespace[2pt]
GPT-5.6 Sol~\citep{openai2026gpt56} & 49.8 & 31.0 & 57.6 & 44.3 & 67.1 & 50.8 & 60.5 & 77.4 & 41.5 & 55.2 \\
GPT-6 Astra~\citep{openai2026gpt6astra} & \textbf{59.6} & \textbf{40.6} & \textbf{68.6} & \textbf{54.6} & \textbf{75.8} & \textbf{63.3} & \textbf{69.9} & \textbf{80.8} & \textbf{48.9} & \textbf{64.6} \\
Gemini 3.1 Pro Preview~\citep{deepmind2026gemini31pro} & 34.8 & 18.1 & 41.8 & 29.9 & 49.4 & 34.5 & 43.6 & 64.5 & 30.7 & 39.8 \\
\midrule[1pt]
\modelgroup{Open-source VLMs} \\
\addlinespace[2pt]
Qwen2.5-VL-72B-Instruct~\citep{bai2025qwen25vl} & 24.7 & 8.2 & 26.1 & 17.2 & 43.4 & 26.8 & 32.8 & 49.3 & 25.4 & 32.2 \\
Qwen3-VL-32B-Instruct~\citep{bai2025qwen3vl} & 29.2 & 11.8 & 32.0 & 21.9 & 48.5 & 30.8 & 37.2 & 58.2 & 29.2 & 36.6 \\
Qwen3-VL-8B-Instruct~\citep{bai2025qwen3vl} & 25.2 & 10.3 & 28.1 & 19.2 & 43.7 & 25.5 & 32.2 & 51.5 & 23.0 & 31.2 \\
InternVL3.5-8B~\citep{wang2025internvl35} & 20.4 & 8.8 & 25.4 & 17.1 & 32.4 & 19.4 & 25.5 & 44.6 & 16.9 & 23.7 \\
MiniCPM-V-4.5-8B~\citep{openbmb2025minicpmv45} & 19.4 & 6.7 & 20.3 & 13.5 & 33.9 & 21.9 & 25.2 & 34.5 & 20.3 & 25.3 \\
VideoLLaMA3-7B~\citep{zhang2025videollama3} & 14.2 & 4.4 & 11.1 & 7.8 & 24.1 & 17.4 & 24.8 & 38.6 & 16.3 & 20.7 \\
Qwen3-VL-4B-Instruct~\citep{bai2025qwen3vl} & 25.9 & 9.9 & 28.2 & 19.1 & 46.3 & 25.2 & 35.0 & 47.8 & 24.0 & 32.8 \\
Molmo2-8B~\citep{clark2026molmo2} & 18.8 & 8.0 & 21.2 & 14.6 & 30.4 & 19.1 & 24.6 & 41.3 & 17.3 & 23.0 \\
LLaVA-OV2-8B~\citep{an2026llavaov2} & 24.7 & 9.8 & 26.7 & 18.2 & 42.0 & 25.8 & 32.6 & 48.1 & 23.9 & 31.2 \\
\midrule
\modelgroup{Open-source video captioning models} \\
\addlinespace[2pt]
video-SALMONN-2+-3B~\citep{tang2025videosalmonn2} & 10.8 & 3.5 & 14.2 & 8.8 & 19.9 & 8.5 & 13.7 & 28.4 & 9.0 & 12.8 \\
CapRL-Video-4B~\citep{yang2026caprlplusplus} & 32.4 & 16.9 & 40.1 & 28.5 & 49.3 & 28.7 & 39.0 & 60.6 & 27.5 & 36.3 \\
OmniCaptioner-IF-3B~\citep{wang2026omnicapif} & 20.0 & 6.6 & 25.7 & 16.1 & 36.0 & 18.0 & 21.9 & 28.8 & 19.5 & 23.8 \\
OmniCaptioner-IF-7B~\citep{wang2026omnicapif} & 23.0 & 10.1 & 25.0 & 17.6 & 38.5 & 24.3 & 26.3 & 40.8 & 24.6 & 28.5 \\
\midrule
\modelgroup{Open-source egocentric video understanding models} \\
\addlinespace[2pt]
EgoThinker-7B~\citep{pei2025egothinker} & 20.8 & 13.7 & 24.1 & 18.9 & 33.7 & 19.0 & 19.1 & 13.1 & 19.5 & 22.8 \\
EgoGPT-7B~\citep{yang2025egolife} & 15.4 & 4.5 & 16.0 & 10.3 & 24.1 & 18.9 & 18.4 & 22.5 & 20.3 & 20.4 \\
RynnBrain-4B~\citep{dang2026rynnbrain} & 13.9 & 10.4 & 13.3 & 11.8 & 21.2 & 12.9 & 15.4 & 15.7 & 14.4 & 15.9 \\
Flora~\citep{leng2026flora} & 37.8 & 18.6 & 46.9 & 32.8 & 51.2 & 41.4 & 37.3 & 59.0 & \textbf{41.3} & 42.9 \\

\midrule
\modelgroup{EgoAction (Ours)} \\
\addlinespace[2pt]
\textbf{EgoAction (LLaVA-OV2-8B)} & \textbf{47.9} & \textbf{32.3} & \textbf{57.0} & \textbf{44.6} & \textbf{67.0} & \textbf{46.2} & \textbf{51.7} & \textbf{63.0} & 39.6 & \textbf{51.2} \\
\textbf{EgoAction (Qwen3-VL-4B)} & 41.8 & 26.6 & 45.9 & 36.3 & 63.4 & 41.5 & 47.6 & 58.2 & 36.8 & 47.4 \\
\bottomrule
\end{tabular}
\end{adjustbox}
\end{table}

\paragraph{Task-specific adaptation improves dense action captioning across backbones.}
EgoAction improves EgoActionScore by 15.9 points on Qwen3-VL-4B~\citep{bai2025qwen3vl} and
23.2 points on LLaVA-OV2-8B~\citep{an2026llavaov2}, with both variants outperforming all evaluated
open-source baselines (Table~\ref{tab:main_results}). Improvements span temporal localization, hand
assignment, and all four semantic fields, with larger gains in the Temporal
component. The OV2 variant also approaches GPT-5.6 Sol~\citep{openai2026gpt56} in Temporal and Hand
scores, while a larger gap remains in Semantic. This pattern suggests that
task-specific training is particularly effective at improving when actions
occur and which hand performs them.

\paragraph{Precise temporal localization remains a central bottleneck.}
GPT-6 Astra reaches an m-tIoU of 40.6, compared with 88.4 for human experts
(Table~\ref{tab:main_results}). Both EgoAction variants also remain well
below human performance on Motion, suggesting that accurately describing
movements such as lifting, rotating, and pressing remains challenging.
Short actions are particularly sensitive to boundary errors, since the same
absolute timing offset accounts for a larger fraction of their duration.
Examining performance across action durations helps identify where temporal
accuracy deteriorates and whether adaptation improves the recovery of brief actions.

\FloatBarrier
\subsection{Performance by Action Duration}
\label{sec:action_duration}

We examine how action duration affects model performance, with particular
attention to actions shorter than one second.
We compare each backbone with its EgoAction variant to assess whether
adaptation improves the description and temporal localization of short actions.
Table~\ref{tab:action_duration} reports Content, Semantic, m-tIoU, and Hand
across duration groups.

\begin{table}[!htbp]
\centering
\footnotesize
\setlength{\tabcolsep}{1.8pt}
\renewcommand{\arraystretch}{1.12}
\caption{\textbf{Performance by ground-truth action duration.}
Per-action components are on a 0--100 scale. Shaded columns highlight actions shorter than one second.
Duration bins (s) are $<0.5$, $[0.5,1)$, $[1,2)$, $[2,5)$, and $\geq5$.
}
\label{tab:action_duration}
\begin{adjustbox}{max width=\linewidth}
\begin{tabular}{@{}l@{\hspace{5pt}}>{\columncolor{blue!5}}r >{\columncolor{blue!5}}r rrr@{\hspace{5pt}}>{\columncolor{blue!5}}r >{\columncolor{blue!5}}r rrr@{\hspace{5pt}}>{\columncolor{blue!5}}r >{\columncolor{blue!5}}r rrr@{\hspace{5pt}}>{\columncolor{blue!5}}r >{\columncolor{blue!5}}r rrr@{}}
\toprule
\multirow{2}{*}{Model} & \multicolumn{5}{c}{\textbf{Content}} & \multicolumn{5}{c}{\textbf{Semantic}} & \multicolumn{5}{c}{\textbf{$\mathrm{m\mbox{-}tIoU}$}} & \multicolumn{5}{c}{\textbf{Hand}} \\
\cmidrule(lr){2-6}\cmidrule(lr){7-11}\cmidrule(lr){12-16}\cmidrule(lr){17-21}
 & $<0.5$ & 0.5--1 & 1--2 & 2--5 & $\geq5$ & $<0.5$ & 0.5--1 & 1--2 & 2--5 & $\geq5$ & $<0.5$ & 0.5--1 & 1--2 & 2--5 & $\geq5$ & $<0.5$ & 0.5--1 & 1--2 & 2--5 & $\geq5$ \\
\midrule
Qwen3-VL-4B & 25.3 & 26.6 & 30.4 & 34.9 & 42.2 & 29.5 & 31.3 & 34.9 & 39.6 & 46.6 & 3.6 & 7.9 & 13.4 & 21.3 & 39.1 & 42.3 & 45.4 & 48.5 & 54.2 & 60.8 \\
\quad + EgoAction\phantom{ SFT} & 35.0 & 41.8 & 48.0 & 53.5 & 59.4 & 39.8 & 47.1 & 53.5 & 59.2 & 65.1 & 18.4 & 26.1 & 30.3 & 40.4 & 56.1 & 54.4 & 63.2 & 70.5 & 77.4 & 83.4 \\
\addlinespace[2pt]
LLaVA-OV2-8B & 23.3 & 25.9 & 30.8 & 36.6 & 43.9 & 26.5 & 29.4 & 34.6 & 40.6 & 48.4 & 3.2 & 7.6 & 13.6 & 21.8 & 40.5 & 36.1 & 40.2 & 46.3 & 53.3 & 63.4 \\
\quad + EgoAction\phantom{ SFT} & 39.3 & 45.8 & 52.0 & 56.1 & 61.5 & 44.1 & 51.1 & 57.4 & 61.4 & 66.5 & 20.3 & 33.5 & 40.3 & 48.9 & 59.3 & 58.8 & 67.2 & 73.9 & 78.2 & 83.0 \\
\addlinespace[2pt]
Qwen3-VL-32B & 28.1 & 30.5 & 36.7 & 41.8 & 50.0 & 31.8 & 34.6 & 40.5 & 45.9 & 54.5 & 4.3 & 9.6 & 15.9 & 25.0 & 45.3 & 42.9 & 47.0 & 52.4 & 58.7 & 69.2 \\
GPT-6 Astra & 51.9 & 61.5 & 69.6 & 72.4 & 76.8 & 55.9 & 65.2 & 73.0 & 75.7 & 79.5 & 23.3 & 42.7 & 55.7 & 62.7 & 68.1 & 67.7 & 76.4 & 83.6 & 86.0 & 88.5 \\
\bottomrule
\end{tabular}
\end{adjustbox}
\end{table}

\paragraph{Brief actions remain difficult to localize precisely.}
Table~\ref{tab:action_duration} shows that performance declines as actions
become shorter across all six models, affecting content recovery, semantics,
temporal overlap, and hand assignment. Temporal localization is particularly
weak: both base models score below 4 in m-tIoU for actions shorter than
0.5 seconds. Task-specific adaptation improves all four components on both
backbones, yet a substantial gap in temporal overlap persists between brief
and longer actions. These results identify precise boundary recovery for
brief actions as a central challenge in dense egocentric action understanding.
\FloatBarrier

\Needspace{8\baselineskip}
\subsection{Ablation Studies}
\label{sec:ablations}
\label{sec:ablation_studies}

\begin{table}[!htbp]
\centering
\scriptsize
\setlength{\tabcolsep}{3pt}
\renewcommand{\arraystretch}{1.12}
\caption{\textbf{Supervised fine-tuning with and without GRPO.}
Performance on EgoActionBench for the Qwen3-VL-4B and LLaVA-OV2-8B backbones.
We report EgoActionScore and its Temporal and Semantic components.
Bold values indicate the higher score within each backbone.
All scores range from 0 to 100.
}
\label{tab:training_stages}
\begin{adjustbox}{max width=\linewidth}
\begin{tabular}{@{}llrrr@{}}
\toprule
Backbone & Training & EgoActionScore & Temporal & Semantic \\
\midrule
\multirow{2}{*}{Qwen3-VL-4B} & SFT & 40.6 & 35.0 & 46.2 \\
 & SFT + GRPO & \textbf{41.8} & \textbf{36.3} & \textbf{47.4} \\
\midrule
\multirow{2}{*}{LLaVA-OV2-8B} & SFT & 46.9 & 43.5 & 50.2 \\
 & SFT + GRPO & \textbf{47.9} & \textbf{44.6} & \textbf{51.2} \\
\bottomrule
\end{tabular}
\end{adjustbox}
\end{table}

\paragraph{GRPO improves performance beyond supervised fine-tuning.}
To assess the contribution of reinforcement learning, we compare the SFT models
with their GRPO-refined counterparts in Table~\ref{tab:training_stages}.
Adding GRPO increases EgoActionScore from 40.6 to 41.8 on Qwen3-VL-4B
and from 46.9 to 47.9 on LLaVA-OV2-8B, with improvements in both Temporal
and Semantic scores. The improvements in both components suggest that our
task-specific reward design strengthens action localization and description
beyond supervised fine-tuning.

\begin{table}[!htbp]
\centering
\scriptsize
\setlength{\tabcolsep}{3pt}
\renewcommand{\arraystretch}{1.12}
\caption{\textbf{Sensitivity to candidate overlap thresholds and frame sampling.}
EgoActionScore on EgoActionBench under different evaluation thresholds and video input settings.
(a) The candidate overlap threshold $\tau_{\mathrm{sem}}$ takes values of 0.3, 0.5, and 0.7, with predictions generated at 2 fps and their semantic judgments held fixed.
(b) We compare sampling at 2, 5, and 10 fps with fixed budgets of 16, 32, and 64 frames per clip, using $\tau_{\mathrm{sem}}=0.5$.
All scores range from 0 to 100.
}
\label{tab:ablation_overall}
\label{tab:temporal_threshold}
\label{tab:input_settings}
\begin{adjustbox}{max width=\linewidth}
\begin{tabular}{@{}lrrrrrrrrr@{}}
\toprule
\multirow{3}{*}{Model} & \multicolumn{3}{c}{(a) Varying $\tau_{\mathrm{sem}}$}
& \multicolumn{6}{c}{(b) Varying video input settings} \\
\cmidrule(lr){2-4}\cmidrule(lr){5-10}
& \multirow{2}{*}{0.3} & \multirow{2}{*}{0.5} & \multirow{2}{*}{0.7}
& \multicolumn{3}{c}{Fixed-rate sampling} & \multicolumn{3}{c}{Fixed-count sampling} \\
\cmidrule(lr){5-7}\cmidrule(lr){8-10}
& & & & 2 fps & 5 fps & 10 fps & 16 frames & 32 frames & 64 frames \\
\midrule
Qwen3-VL-4B & 26.4 & 25.9 & 24.8 & 25.9 & 27.6 & 27.6 & 22.2 & 25.4 & 26.8 \\
Qwen3-VL-32B & 29.9 & 29.2 & 27.9 & 29.2 & 31.6 & 31.3 & 27.0 & 29.1 & 30.6 \\
LLaVA-OV2-8B & 25.3 & 24.7 & 23.7 & 24.7 & 26.5 & 26.3 & 21.9 & 23.7 & 25.4 \\
GPT-6 Astra & 60.9 & 59.6 & 55.6 & 59.6 & 62.9 & 64.3 & 51.9 & 58.3 & 61.5 \\
\midrule
EgoAction (Qwen3-VL-4B) & 43.1 & 41.8 & 38.4 & 41.8 & 49.8 & 50.7 & 31.7 & 39.0 & 46.0 \\
EgoAction (LLaVA-OV2-8B) & 49.5 & 47.9 & 43.5 & 47.9 & 52.8 & 53.4 & 38.6 & 44.6 & 50.4 \\
\bottomrule
\end{tabular}
\end{adjustbox}
\end{table}

\paragraph{Model rankings remain stable across overlap thresholds.}
To assess sensitivity to candidate filtering, we vary
$\tau_{\mathrm{sem}}\in\{0.3,0.5,0.7\}$ while keeping predictions and
pairwise semantic judgments fixed. Stricter filtering lowers absolute scores
but preserves the ranking of all six models
(Table~\ref{tab:ablation_overall}). Both EgoAction variants retain substantial
gains over their backbones, including improvements of 13.6 and 19.8 points
at the strictest threshold. The model comparisons and benefits of adaptation
therefore remain consistent across the tested thresholds.

\paragraph{More input frames improve EgoAction performance.}
Sparse video sampling may miss brief actions and changes at their boundaries.
We vary the frame rate and frame budget to examine sensitivity to sampling
and whether EgoAction maintains its advantage across input settings
(Table~\ref{tab:ablation_overall}(b) and Figure~\ref{fig:ablations}(e--f)).
All six models improve when the sampling rate increases from 2 to 5 fps.
Further increasing it to 10 fps yields limited additional gains.
Under fixed-count sampling, all models improve with a larger frame budget.
Both EgoAction variants outperform their backbones under every tested setting.

\begingroup
\setlength{\intextsep}{8pt plus 2pt minus 2pt}
\begin{figure}[htbp]
\centering
\setlength{\abovecaptionskip}{4pt}
\includegraphics[width=0.75\linewidth]{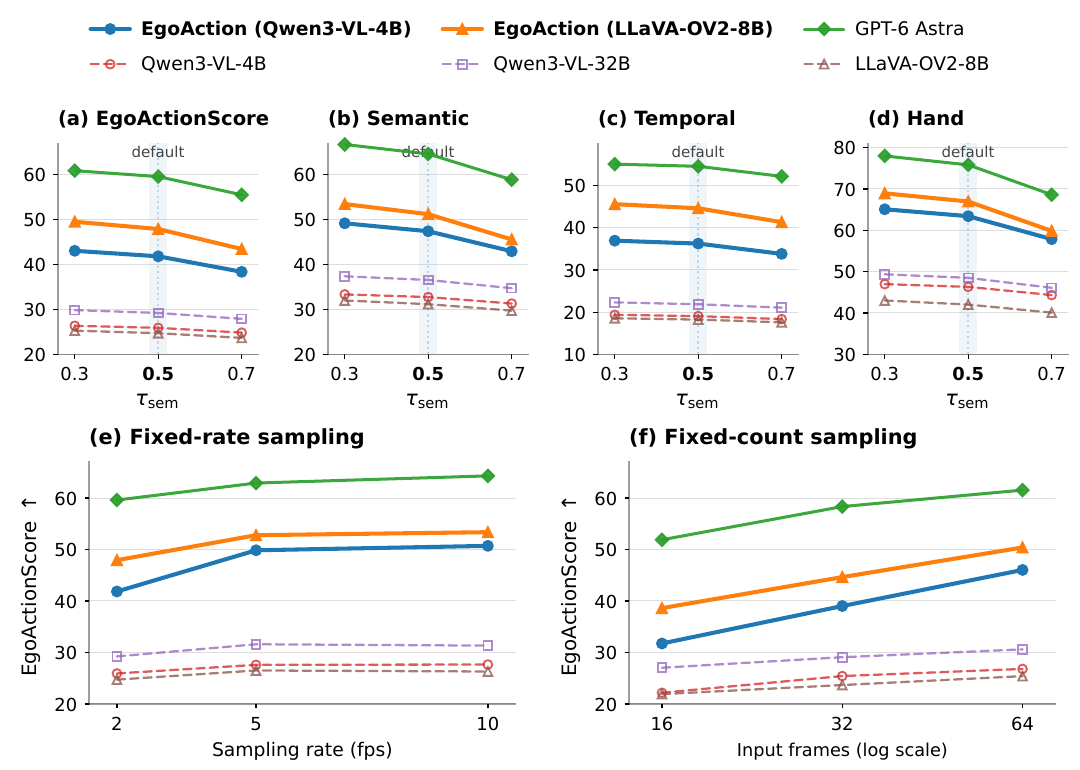}
\caption{\textbf{Metric sensitivity to candidate overlap thresholds and frame sampling.}
Panels (a) to (d) show EgoActionScore, Semantic, Temporal, and Hand as the candidate overlap threshold varies, using predictions generated at 2 fps.
The shaded bands mark the default threshold of $\tau_{\mathrm{sem}}=0.5$.
Panels (e) and (f) show EgoActionScore under fixed-rate and fixed-count sampling, with $\tau_{\mathrm{sem}}=0.5$.
Panel (f) uses a logarithmic frame-count axis.
All scores range from 0 to 100; vertical axis ranges vary across panels.}
\label{fig:ablations}
\end{figure}
\endgroup

\FloatBarrier

\Needspace{10\baselineskip}
\section{Conclusion}
\label{sec:conclusion}

We introduced EgoActionBench, a benchmark for dense action captioning in
egocentric video. It combines atomic action annotations with an evaluation
protocol for action content, temporal precision, and hand association.
Our experiments show that current models remain well below human performance
on this task. Supervised fine-tuning followed by GRPO improves both backbones,
but accurately locating brief actions and describing their movements remain
challenging. We hope EgoActionBench will encourage the development of models
that capture the action details needed to learn from human demonstrations.

\begingroup
\raggedright
\setlength{\bibsep}{2pt plus 1pt minus 0.5pt}
\clearpage
\bibliography{main}
\bibliographystyle{assets/plainnat}
\endgroup

\clearpage
\appendix
\section{Additional Experimental Details}
\label{sec:implementation_details}

\paragraph{Adaptation data.}
The 5,354 automatically annotated episodes comprise 5,146 training examples
and 208 validation examples. Source-video identifiers separate this corpus
from EgoActionBench. The benchmark annotations undergo human verification;
the adaptation labels are generated and refined automatically.

\paragraph{Supervised adaptation.} We fine-tune the language models and multimodal connectors with frozen visual encoders, using a learning rate of $10^{-5}$, a cosine schedule, and an effective batch size of 32.

\paragraph{Reinforcement learning.}
Both backbones are initialized from their SFT models. We update the
language-model parameters while keeping the visual encoders and multimodal
connectors frozen. The initial SFT policies serve as fixed references for
KL regularization. Training uses the Dr.~GRPO loss~\citep{liu2025understanding}.
Each candidate's reward is centered by its video-group mean and scaled by
the reward standard deviation across the generation batch. A fixed
Qwen3-32B judge assesses predicted captions against the training captions
and their extracted semantic fields. The reward components $S$, $I$, and
$F$ are computed per candidate output on a $[0,1]$ scale.
All training runs use random seed 42, and each GRPO rollout batch is used
for a single policy update.
Table~\ref{tab:training_configuration} summarizes the key hyperparameters
for SFT and GRPO.

\begin{table}[!htbp]
\centering
\small
\setlength{\tabcolsep}{6pt}
\renewcommand{\arraystretch}{1.10}
\caption{\textbf{Key optimization and sampling hyperparameters for SFT and GRPO.}
Values apply to both backbones. Batch size counts supervised examples for SFT
and generated responses for GRPO. Each GRPO batch contains six videos with
$G=4$ responses per video. A dash indicates that the parameter is not applicable.}
\label{tab:training_configuration}
\begin{tabularx}{\linewidth}{@{}l>{\raggedright\arraybackslash}X>{\raggedright\arraybackslash}X@{}}
\toprule
Hyperparameter & SFT & GRPO \\
\midrule
Optimizer & AdamW & AdamW \\
Learning rate & $1\times10^{-5}$ & $5\times10^{-7}$ \\
Learning rate scheduler & Cosine & Cosine \\
Learning rate warmup & 5\% of training updates & 10 updates \\
Training batch size & 32 & 24 \\
Gradient norm clipping & 1.0 & 1.0 \\
KL coefficient $\beta$ & \textemdash & 0.04 \\
Group size $G$ & \textemdash & 4 \\
Sampling temperature & \textemdash & 0.8 \\
Top-$p$ & \textemdash & 0.95 \\
\bottomrule
\end{tabularx}
\end{table}

\paragraph{Semantic evaluation.}
Qwen3-32B is the text-only semantic judge, with temperature zero.
It receives the reference caption, its structured semantic fields, and the
predicted caption. Every temporally eligible candidate is evaluated.
The default candidate-overlap threshold is $\tau_{\mathrm{sem}}=0.5$;
selection follows \cref{sec:benchmark_metrics}.
When content score, hand-label agreement, and temporal IoU are all equal,
we resolve ties by earlier start time, followed by earlier end time.
Any remaining ties are resolved deterministically.
Semantic fidelity and matched temporal overlap are averaged over ground-truth
actions, including zero scores for actions without positive-content candidates.
F1 uses dataset-level counts from one-to-one matching and retains all
submitted predictions in its denominator. Temporal averages matched overlap
and F1; EgoActionScore averages Temporal and Semantic.

\paragraph{Qwen evaluation settings.}
We evaluate EgoAction (Qwen3-VL-4B) with video sampled at 2 fps and the
inference prompt in Figure~\ref{fig:inference_prompt}.
Both Qwen3-VL-4B variants use temperature zero, random seed 42, and reasoning
disabled.

\clearpage
\begingroup
\setlength{\parskip}{2pt}
\setlength{\intextsep}{1.5pt}
\setlength{\textfloatsep}{2pt}
\titlespacing*{\subsection}{0pt}{4pt}{2pt}
\captionsetup{skip=3pt}
\section{Robustness to the Semantic Judge}
\label{app:judge_robustness}

\subsection{Agreement with Human Reviewers}

Two human reviewers independently evaluated a 400-pair calibration set
containing 1,378 required field items. Across 1,377 jointly decidable items,
Qwen3-32B agrees with Reviewer 1 and Reviewer 2 on 88.89\% and 89.03\%
of binary credit decisions, with Cohen's $\kappa$ of 0.742 and 0.746.
Exact label agreement over all 1,378 items is 65.24\% and
67.85\%; most remaining disagreements distinguish missing information
from contradictions, both of which receive zero credit.

\begin{figure}[!ht]
\centering
\includegraphics[width=0.77\linewidth]{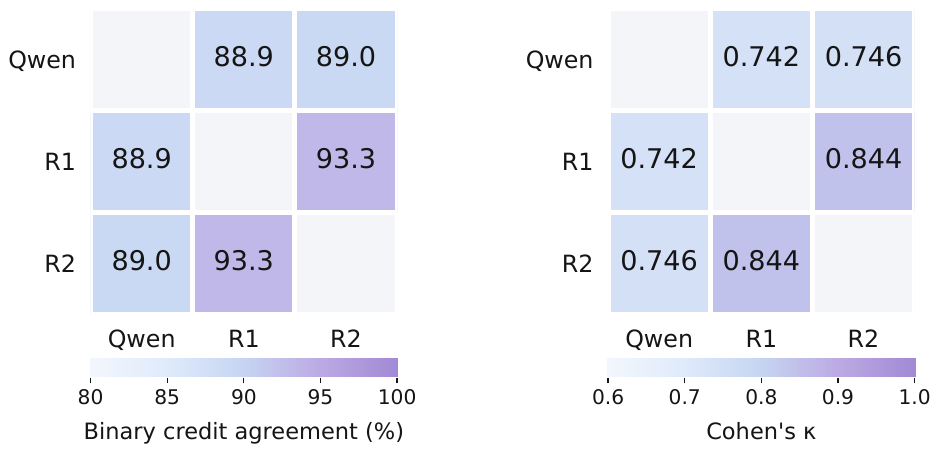}
\caption{\textbf{Agreement with human reviewers on the 400-pair calibration set.} Percentage agreement
(left) and Cohen's $\kappa$ (right) on 1,377 semantic judgments, based on
whether each requirement is satisfied. Qwen denotes Qwen3-32B;
R1 and R2 denote the two reviewers.}
\label{fig:judge_credit_agreement}
\end{figure}
\FloatBarrier

\subsection{Stability of Model Comparisons}

We rescore fixed predictions from six model configurations on 99 clips
and the same 2,506 ground-truth actions under both judges.
The comparison includes the SFT versions of both EgoAction variants.
Ground truth and candidate sets are held fixed, while the selected prediction
and one-to-one action matching are recomputed for each judge.

\paragraph{Model comparisons remain stable across semantic judges.}
Qwen3-32B and GPT-5.6 Sol produce identical model rankings for EgoActionScore
and Semantic, with Spearman's $\rho=1.0$ for both. Both judges also preserve
the gains from supervised adaptation on the two backbones, supporting the
stability of these comparisons under the tested judge change.

\begin{figure}[!ht]
\centering
\includegraphics[width=\linewidth]{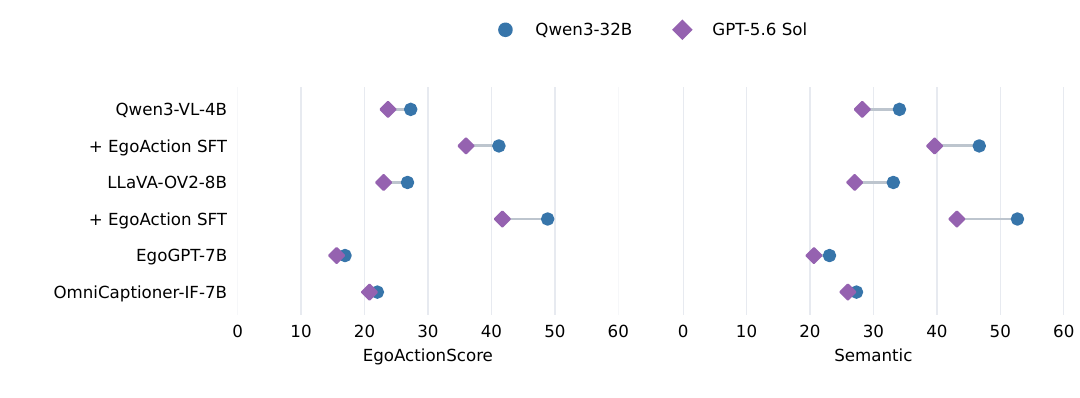}
\caption{\textbf{Model scores under two semantic judges.} Blue circles
and purple diamonds show Qwen3-32B and GPT-5.6 Sol judgments of fixed
predictions on the same 99 clips and 2,506 ground-truth actions.
Both panels share the same axis scale.}
\label{fig:judge_paired_scores}
\end{figure}
\FloatBarrier

\clearpage
\endgroup

\clearpage
\section{Additional Dataset Details}
\label{app:data_collection}

\subsection{Video preparation and temporal coordinates}
\label{app:video_preparation}

The benchmark videos retain source-dependent frame rates and resolutions,
as summarized in \cref{tab:dataset_video_specifications}.
For EgoLife, we use the $960\times960$ central clean-view crop of the
$1280\times1280$ source view. This crop removes the outer region containing
participant and timestamp overlays. Human annotation uses the clean video,
while hand-highlighted videos support the generation of annotation drafts.

\begin{table}[htbp]
\centering
\small
\setlength{\tabcolsep}{6pt}
\renewcommand{\arraystretch}{1.12}
\caption{\textbf{Video specifications by source.} Frame rates and resolutions
of the benchmark episodes.}
\label{tab:dataset_video_specifications}
\begin{tabularx}{\linewidth}{@{}lrr>{\raggedright\arraybackslash}X@{}}
\toprule
Source & Episodes & Frame rate & Video resolution \\
\midrule
Ego4D & 451 & 30 fps & $480\times360$, $640\times360$, $960\times540$, $960\times720$, or $960\times1280$ \\
EgoLife & 485 & 20 fps & $960\times960$ \\
EPIC-KITCHENS-100 & 471 & 50 fps & $1920\times1080$ \\
\bottomrule
\end{tabularx}
\end{table}

We express action boundaries in seconds relative to the first decoded frame
of each episode. For a constant-frame-rate video at $f$ frames per second,
frame $i$ corresponds to $t_i=i/f$, using zero-based indexing.
An action covering frames $i$ through $j-1$ occupies the half-open interval
$[i/f,j/f)$, with duration $(j-i)/f$.
A one-frame boundary shift therefore corresponds to approximately
$33.3$\,ms for Ego4D, $50$\,ms for EgoLife, and $20$\,ms for
EPIC-KITCHENS-100. Annotators inspect native decoded frames around action
transitions to refine the start and end boundaries.

\subsection{Activity coverage and statistics}
\label{app:activity_statistics}

The episodes are organized into 17 scenes across six activity groups.
Kitchen and dining includes cooking, kitchen cleanup, and dining activities.
Household and daily life covers household tasks, laundry, shopping, personal
care, and generic object handling. Construction and repair includes building
work, woodwork, and mechanical repair. The remaining groups cover gardening
and outdoor activities, crafts and laboratory work, and desk, device, and
leisure activities.

We assign initial scene labels from episode task summaries using keyword
rules and refine ambiguous assignments through review. Each episode receives
a scene label, which determines its activity group. This categorization is
used to summarize the activity coverage of the selected data
(\cref{tab:dataset_activity_statistics}).

\begin{table}[htbp]
\centering
\small
\setlength{\tabcolsep}{10pt}
\renewcommand{\arraystretch}{1.10}
\caption{\textbf{Activity coverage of EgoActionBench.} Episode and action-caption
counts across the six activity groups.}
\label{tab:dataset_activity_statistics}
\begin{tabular}{lrr}
\toprule
Activity group & Episodes & Action captions \\
\midrule
Kitchen and dining & 771 & 20,205 \\
Household and daily life & 230 & 6,783 \\
Construction and repair & 185 & 6,832 \\
Gardening and outdoor & 28 & 894 \\
Crafts and laboratory & 103 & 3,812 \\
Desk, devices, and leisure & 90 & 2,606 \\
\midrule
\textbf{Total} & \textbf{1,407} & \textbf{41,132} \\
\bottomrule
\end{tabular}
\end{table}

Kitchen and dining accounts for the largest share of the dataset.
Household activities and construction and repair provide substantial
additional coverage, while the remaining groups extend the benchmark to
outdoor work, crafts, laboratory procedures, and interactions with books
and devices.

Action statistics are computed across the Left, Right, and Both tracks.
Each annotated interval contributes one caption. A collaborative action and
its annotated single-hand roles therefore contribute separate records.
Non-action entries, including idle intervals, hand invisibility, and the
absence of bimanual collaboration, are excluded from action counts.
Total video duration counts each episode once, while the action-duration
distribution includes individual intervals from all three tracks.

\FloatBarrier

\clearpage
\begingroup
\setlength{\parskip}{3pt}
\setlength{\intextsep}{6pt}
\setlength{\textfloatsep}{6pt}
\titlespacing*{\subsection}{0pt}{7pt}{3pt}
\captionsetup{skip=4pt}
\section{Human Annotation Protocol}
\label{app:human_annotation}

\begin{minipage}{\linewidth}
\centering
\includegraphics[width=\linewidth]{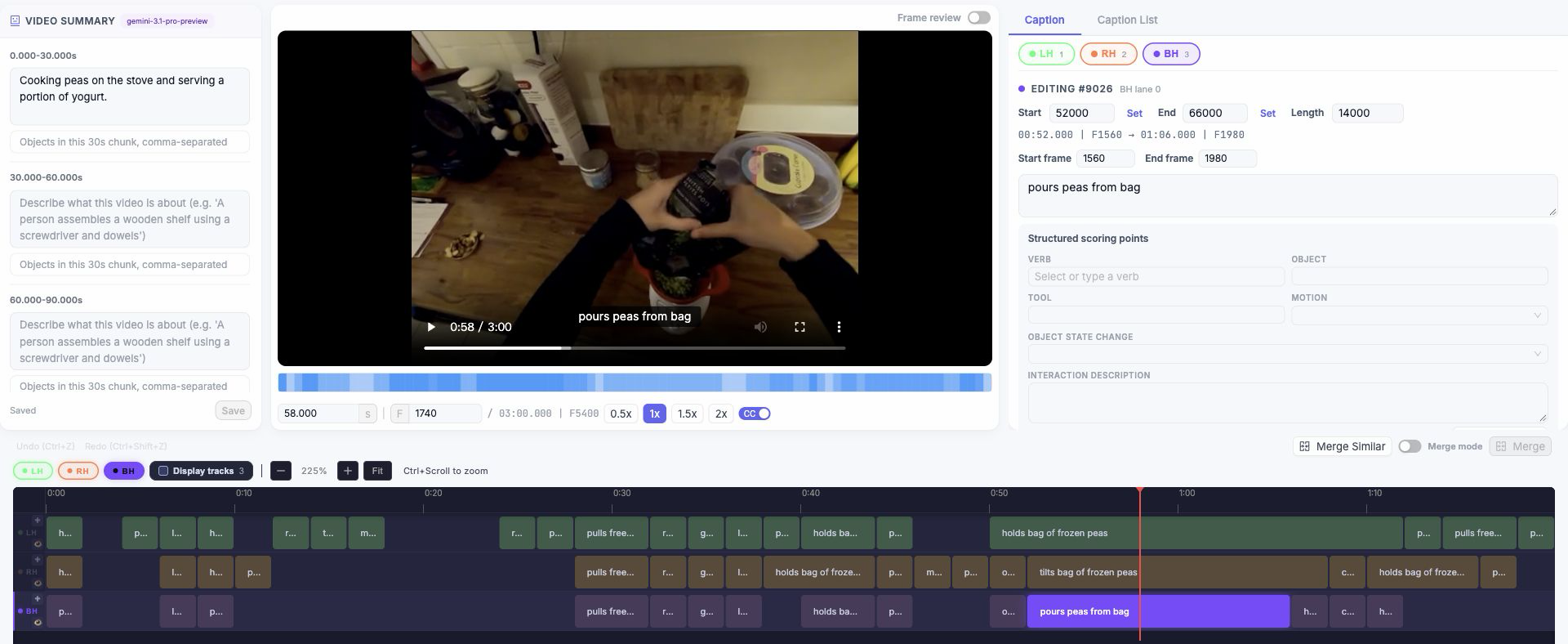}
\captionof{figure}{\textbf{Annotation interface.} The interface displays
 the source video and summary alongside a caption editor and aligned
Left (LH), Right (RH), and Both (BH) timelines. The selected segment
 describes pouring peas from a bag; its caption, timestamps, and frame
indices appear in the editor.}
\label{fig:annotation_interface}
\end{minipage}

\subsection{Annotators and Responsibilities}

An external team of ten annotators constructed the benchmark annotations
in English. Each video was assigned to two team members: one revised the
model-generated drafts against the source video, and the other checked
the revisions for errors and omissions. Three additional human reviewers
performed final quality control. A separate group of three annotators
conducted the human-baseline evaluation.

\subsection{Training and Annotation Guidelines}

Training materials included written guidelines, a pilot annotation batch,
and case-specific feedback. They covered interface operation, action
granularity, hand assignment, bimanual collaboration, and temporal
boundaries. Pilot feedback clarified recurring errors and ambiguous
cases. Annotators followed the task definition, describing individual
action phases and distinguishing each hand's role from bimanual
collaboration.

\subsection{Quality Control and Final Verification}

After annotation and second-person review, agentic quality control
proposed revisions supported by frame references and explanations.
Human reviewers checked these proposals against the source videos,
resolved ambiguities, and approved the annotations or returned them for
correction and further QC.

\subsection{Annotation Interface}

The annotation software provides video playback, a caption editor, and
aligned Left, Right, and Both timelines
(\cref{fig:annotation_interface}). Selecting a segment opens its description
and boundaries for editing. Annotators can inspect individual frames,
revise captions, and adjust timestamps or frame indices while comparing
individual hand roles with bimanual actions.

\FloatBarrier
\clearpage
\endgroup

\clearpage
\begingroup
\setlength{\parskip}{3pt}
\setlength{\intextsep}{6pt}
\captionsetup{skip=4pt}
\titlespacing*{\section}{0pt}{0pt}{6pt}
\titlespacing*{\subsection}{0pt}{6pt}{4pt}
\section{Prompt Templates}
\label{app:prompts}

\subsection{Inference Prompt}
\label{app:inference_prompt}

The inference prompt takes a complete video clip as input and requests
timestamped action captions for the Left, Right, and Both tracks
(Figure~\ref{fig:inference_prompt}).

\begin{figure}[!ht]
\centering
\includegraphics[width=\linewidth]{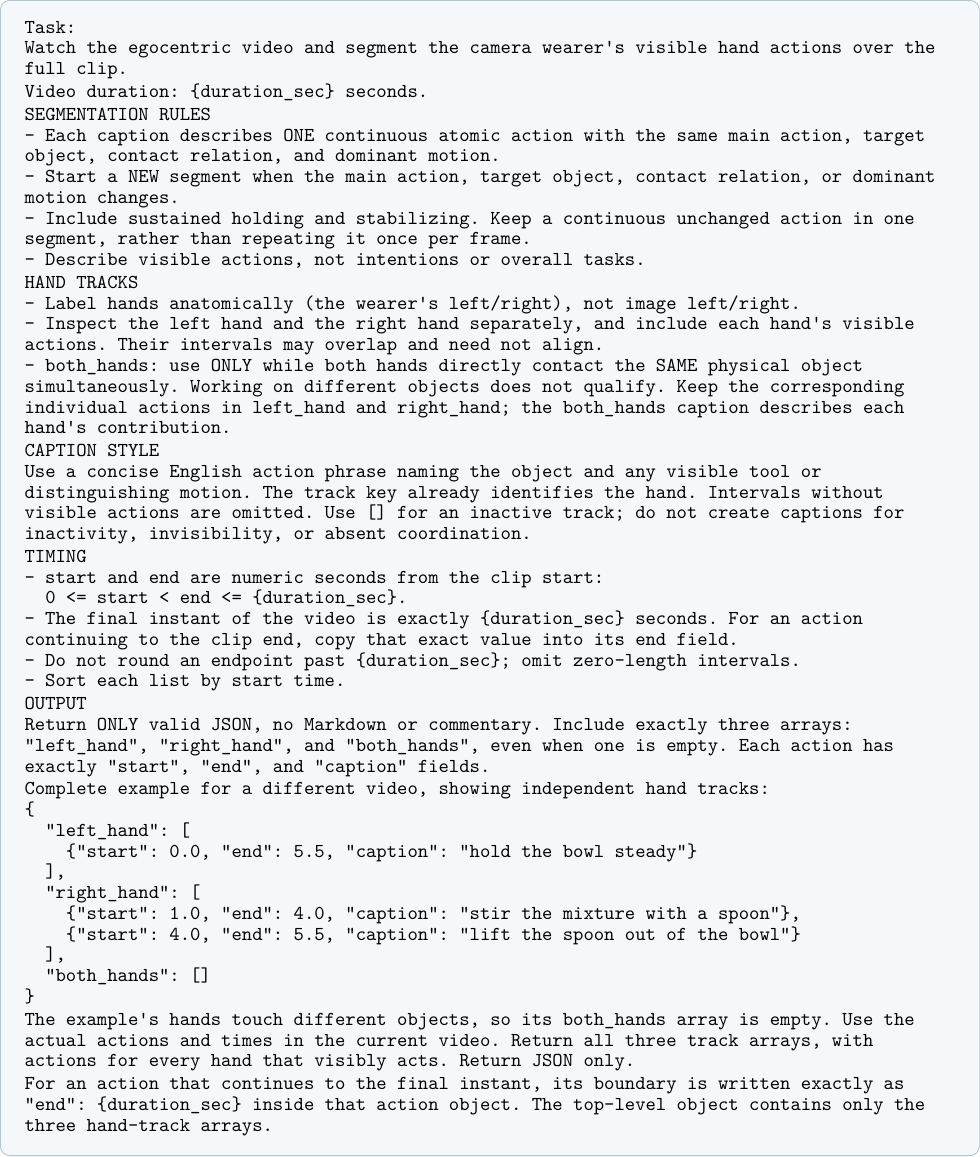}
\caption{\textbf{Inference prompt for dense per-hand action captioning.}
The placeholder \texttt{\{duration\_sec\}} is replaced with the duration of each clip.}
\label{fig:inference_prompt}
\end{figure}
\FloatBarrier
\clearpage
\endgroup

\clearpage
\begingroup
\setlength{\parskip}{3pt}
\setlength{\intextsep}{6pt}
\captionsetup{skip=4pt}
\titlespacing*{\subsection}{0pt}{6pt}{4pt}
\subsection{Prompts for Dataset Construction}
\label{app:construction_prompts}

\looseness=1
The sliding-window template generates local caption proposals
(Figure~\ref{fig:caption_generation_prompt}). The merging instructions
consolidate adjacent proposals describing the same continuous action before
temporal boundary refinement (Figure~\ref{fig:temporal_merging_prompt}).\par

\begin{figure}[!ht]
\centering
\includegraphics[width=\linewidth]{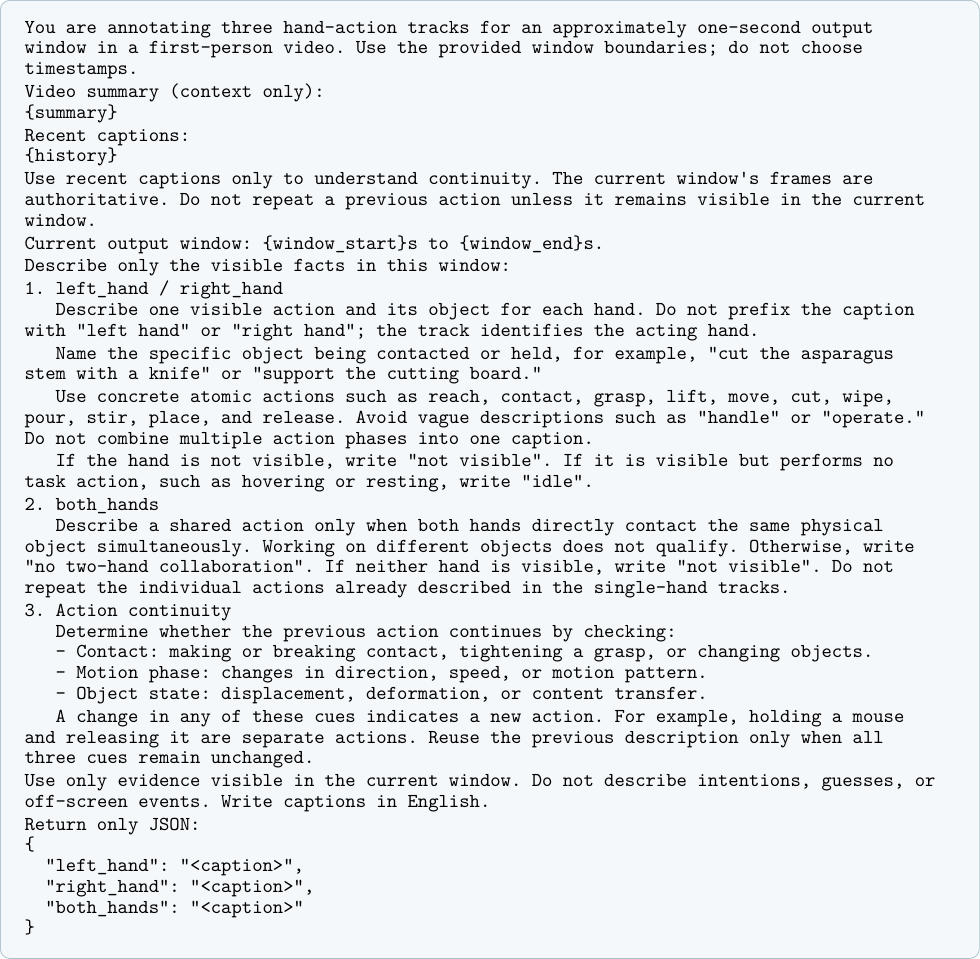}
\caption{\textbf{Template for sliding-window caption generation.}
The inputs include a video summary, recent captions, and the current output
window. The model returns captions for the Left, Right, and Both tracks.}
\label{fig:caption_generation_prompt}
\end{figure}
\FloatBarrier
\clearpage

\begin{figure}[!ht]
\centering
\includegraphics[width=\linewidth]{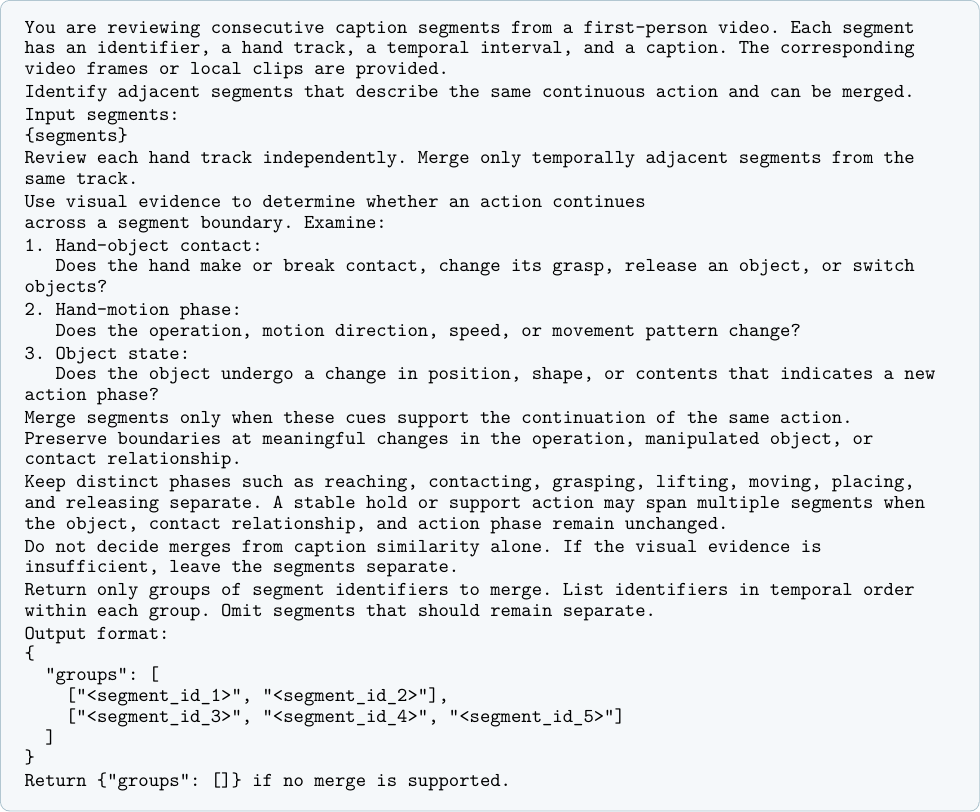}
\caption{\textbf{Instructions for temporal merging.}
Adjacent segments are reviewed within each hand track using visual evidence
of hand-object contact, hand-motion phases, and object-state changes.}
\label{fig:temporal_merging_prompt}
\end{figure}
\FloatBarrier
\clearpage
\endgroup

\section{Supplementary Analysis of Evaluation Metrics}
\label{sec:discussion}

\begin{figure}[!htbp]
\centering
\includegraphics[width=\linewidth]{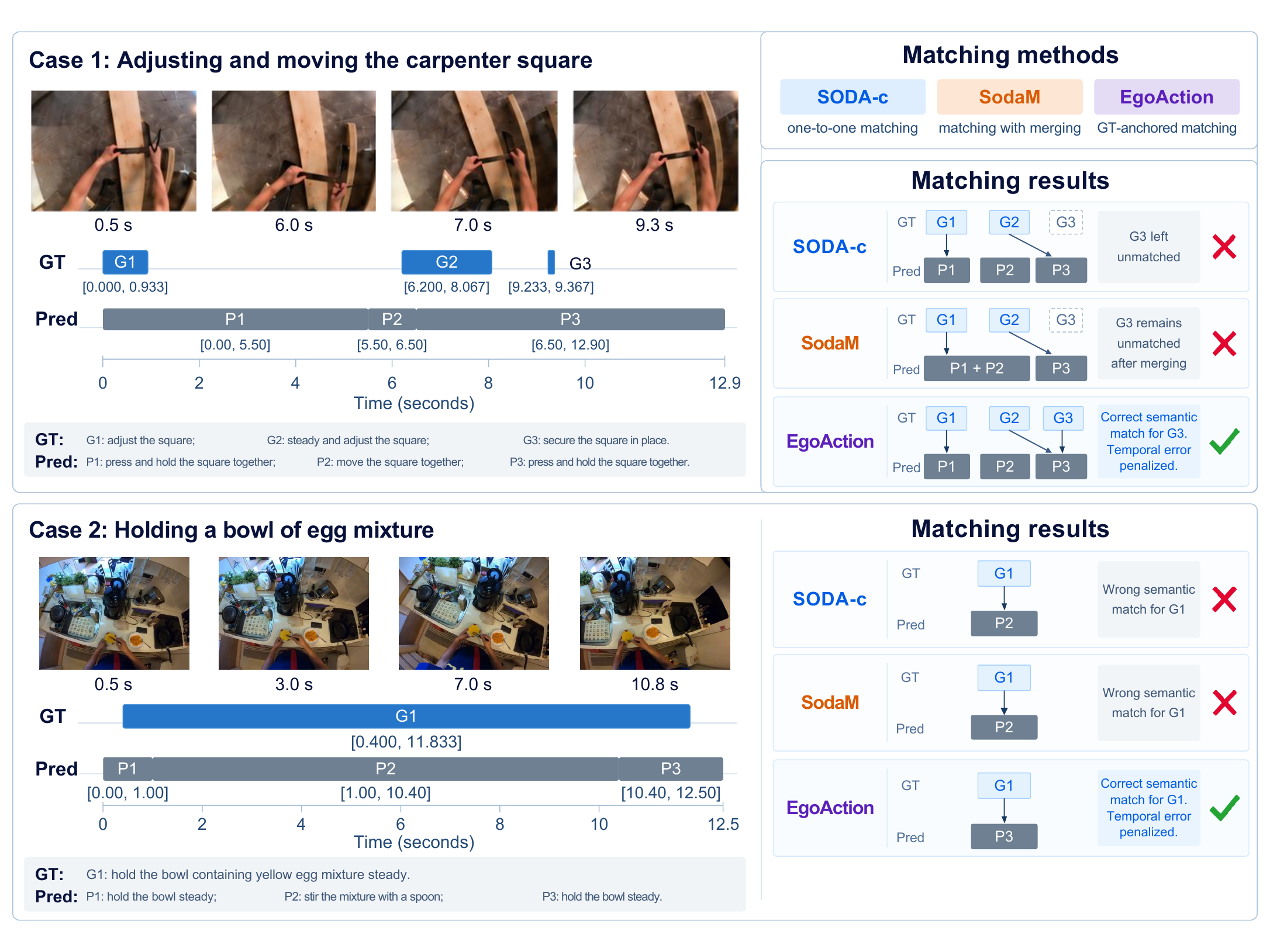}
\caption{\textbf{Qualitative Comparison of Evaluation Metrics.}
We compare SODA-c~\citep{fujita2020soda},
SodaM~\citep{yao2026timechatcaptioner}, and our evaluation protocol.
The examples illustrate how our protocol preserves action-level content
matches while evaluating temporal overlap on the original predictions.}
\label{fig:matching_cases}
\end{figure}

\paragraph{Preserving semantic matches while penalizing segmentation errors.}
Figure~\ref{fig:matching_cases} shows how action-centered matching preserves
semantic credit for correctly described actions. In the carpenter-square case,
SODA-c's one-to-one matching leaves G3 unmatched after assigning P3 to G2.
SodaM allows multiple predictions to be merged for one GT, but still cannot
reuse P3 for G3. Our protocol evaluates each GT action independently,
preserving semantic credit for the content P3 correctly describes in both
references. Temporal IoU penalizes its inaccurate boundaries, while a separate
one-to-one F1 score penalizes under-segmentation. For the bowl-holding action,
P2 has greater temporal overlap with G1 but describes stirring. This overlap
outweighs its lower METEOR similarity in SODA-c, while SodaM determines matches
from temporal overlap alone. By prioritizing action content, our protocol
selects P3, which correctly describes holding the bowl. Its limited overlap
with G1 remains penalized by temporal IoU.

\clearpage
\begingroup
\setlength{\intextsep}{6pt}
\captionsetup{skip=4pt,justification=raggedright,singlelinecheck=false}
\section{Dataset Examples}
\label{app:dataset_examples}

Figures~\ref{fig:dataset_example_ego4d}--\ref{fig:dataset_example_epic}
illustrate the temporal granularity and hand assignments of annotations across
the three data sources. In the Ego4D example, the left hand maintains its grip
on the steering wheel while the right hand handles a writing board. Separate
intervals distinguish reaching, grasping, and lifting from the longer holding
action. The EgoLife example shows a transition from one-handed to two-handed
manipulation of a vase. In the EPIC-KITCHENS-100 example, joint shaking of a
cake mold is followed by separate actions involving a mixing bowl and a pot.
Each example shows selected frames and complete annotations on the Left,
Right, and Both tracks.
All timestamps are measured in seconds from the start of the clip.

\begin{figure}[!ht]
\centering
\includegraphics[width=\linewidth]{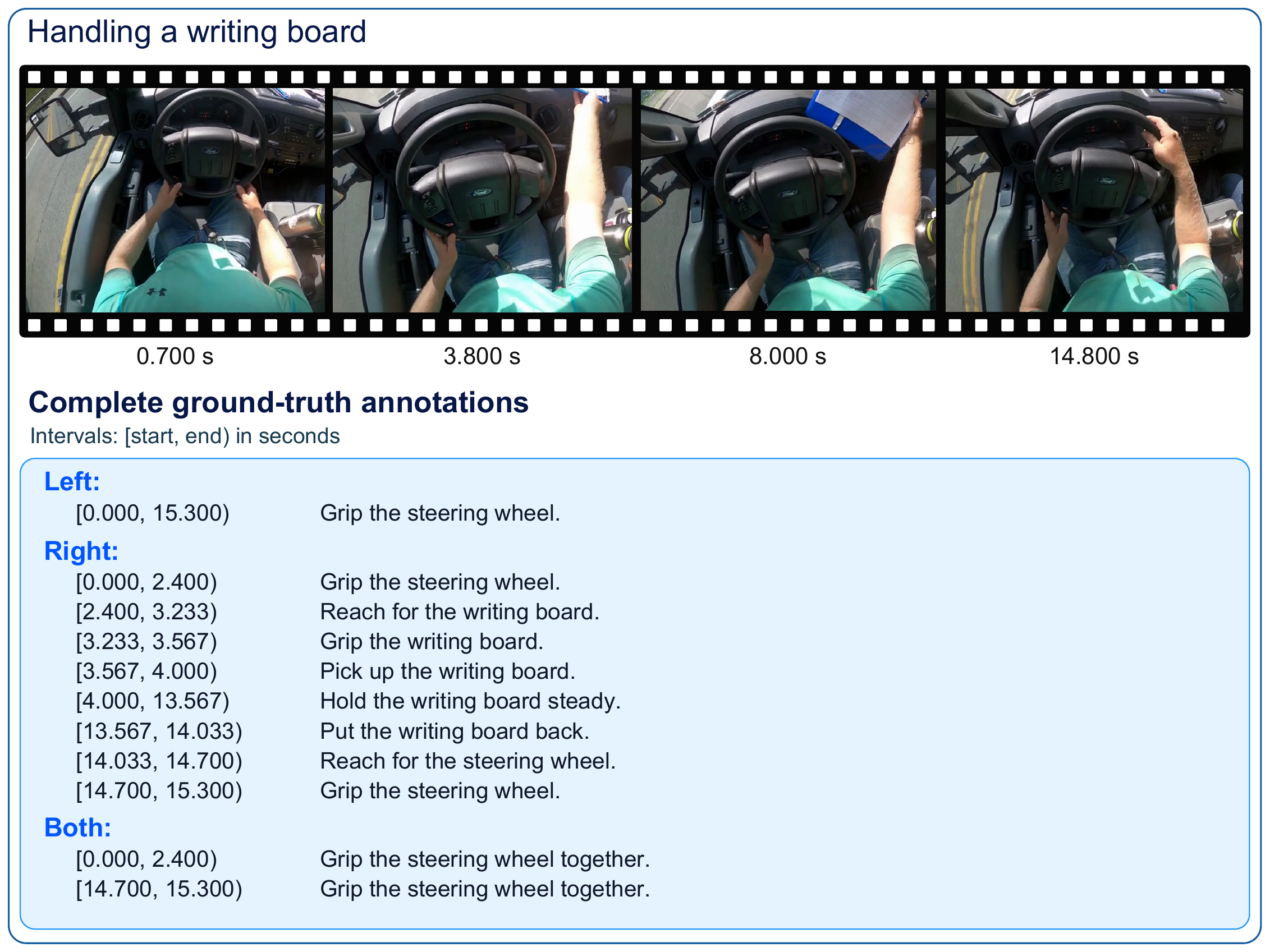}
\caption{\textbf{Ego4D example.} Selected frames and complete ground-truth annotations for handling a writing~board.}
\label{fig:dataset_example_ego4d}
\end{figure}
\FloatBarrier

\clearpage
\begin{figure}[!ht]
\centering
\includegraphics[width=0.82\linewidth]{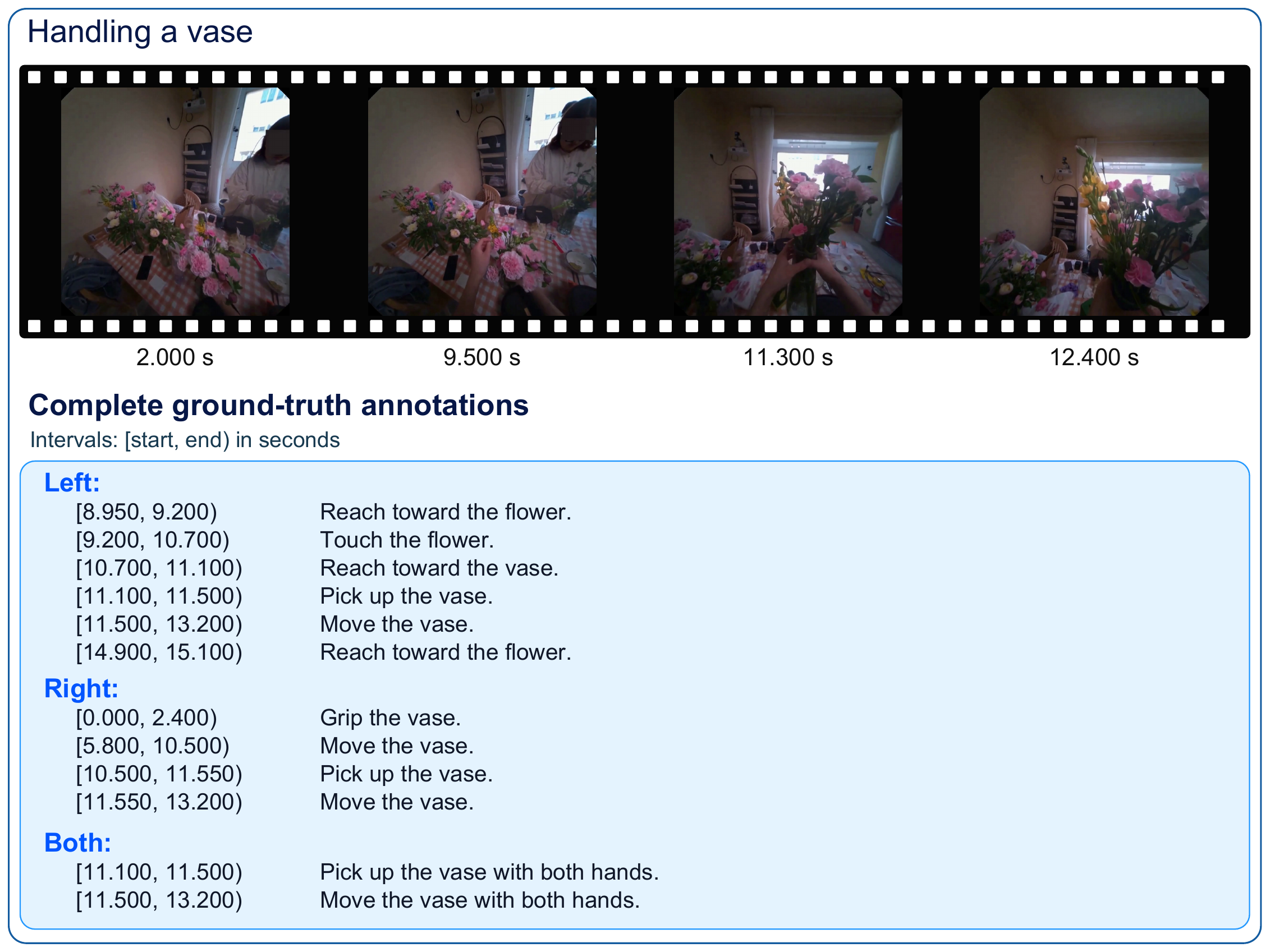}
\caption{\textbf{EgoLife example.} Selected frames and complete ground-truth annotations for handling a~vase.}
\label{fig:dataset_example_egolife}
\par\vspace{2pt}
\includegraphics[width=0.82\linewidth]{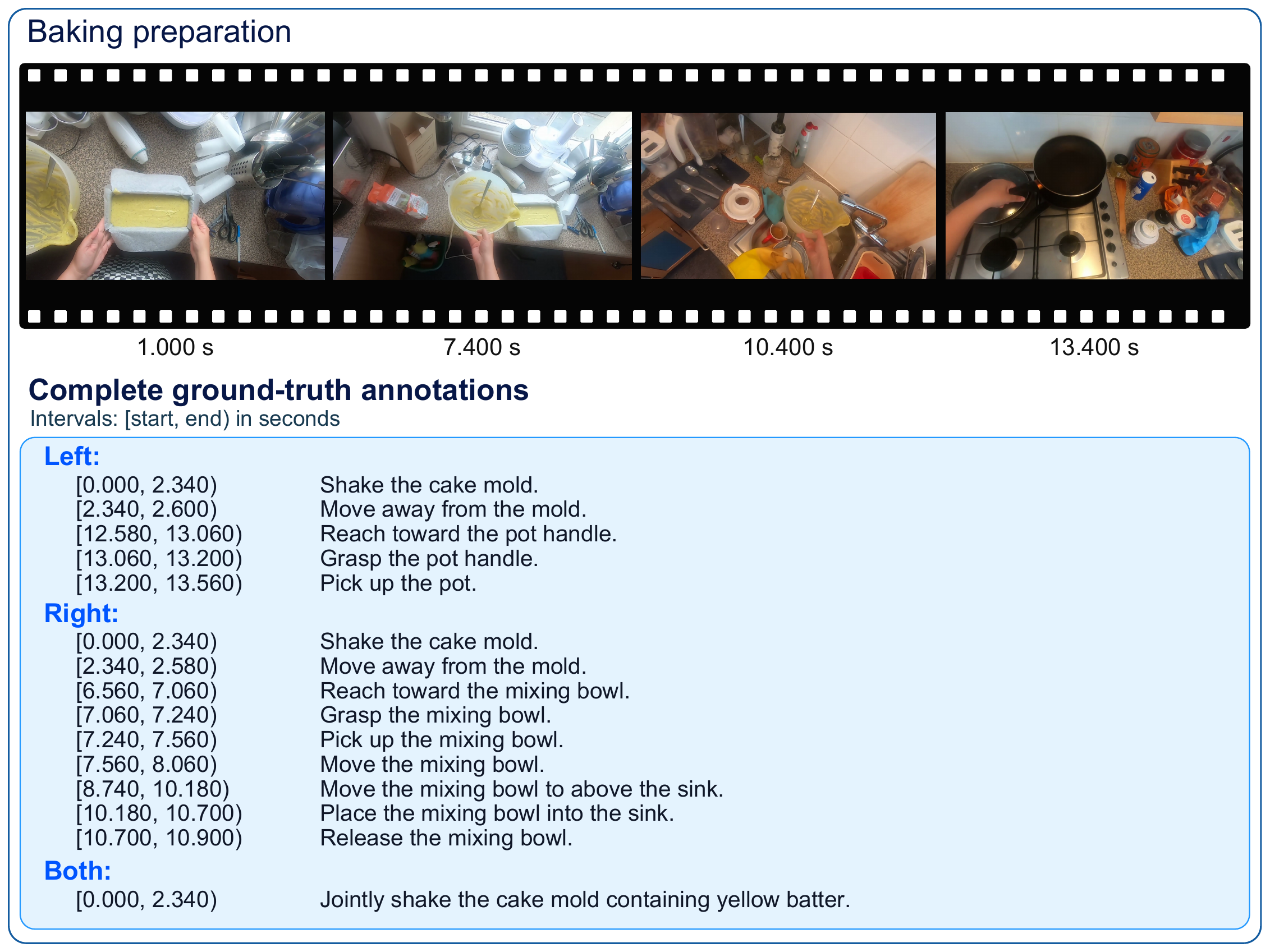}
\caption{\textbf{EPIC-KITCHENS-100 example.} Selected frames and complete ground-truth annotations \mbox{for baking preparation.}}
\label{fig:dataset_example_epic}
\end{figure}
\FloatBarrier
\endgroup

\clearpage
\section{Results by Data Source}
\label{app:source_results}

Tables~\ref{tab:source_ego4d}, \ref{tab:source_egolife},
and~\ref{tab:source_epic_kitchens_100} report
results separately for the Ego4D, EgoLife, and EPIC-KITCHENS-100 subsets.
We use the same models, metrics, and aggregation rules as
Table~\ref{tab:main_results}, with all scores computed within each source.
Bold indicates the best model result in each column for that source;
the human reference is excluded from this comparison.

We compare results by data source to examine whether the main findings
hold across subsets. GPT-6 Astra achieves the highest model scores on
all reported metrics. The LLaVA-OV2 variant achieves the highest
EgoActionScore among open-source baselines on all three sources.
Both EgoAction variants also achieve higher Temporal and Semantic scores
than their respective backbones.
However, GPT-6 Astra remains well below human performance in m-tIoU
across all three sources. Precise temporal localization therefore
remains a common challenge.

\begin{table}[!htbp]
\centering
\scriptsize
\setlength{\tabcolsep}{4pt}
\renewcommand{\arraystretch}{1.12}
\newcommand{\modelgroup}[1]{\multicolumn{11}{@{}l}{\textit{#1}}}
\caption{\textbf{Dense action captioning on the Ego4D subset of EgoActionBench.}
All scores range from 0 to 100; higher is better.
Temporal Overall is the mean of m-tIoU and micro-averaged F1.
Semantic Overall averages the applicable Hand, Verb, Objects, Tools, and Motion scores within each ground-truth action, then averages over ground-truth actions.
EgoActionScore is the mean of Temporal Overall and Semantic Overall.
Parentheses identify the EgoAction backbone.
Best model scores are bold.}
\label{tab:source_ego4d}
\begin{adjustbox}{max width=\linewidth}
\begin{tabular}{@{}l r @{\hspace{7pt}} rrr @{\hspace{7pt}} rrrrrr@{}}
\toprule
\multirow{2}{*}{\textbf{Model}}
& \multirow{2}{*}{\makecell{\textbf{EgoAction}\\\textbf{Score}}}
& \multicolumn{3}{c}{Temporal} & \multicolumn{6}{c}{Semantic} \\
\cmidrule(lr){3-5}\cmidrule(lr){6-11}
& & m-tIoU & F1 & Overall
& Hand & Verb & Objects & Tools & Motion & Overall \\
\midrule
\modelgroup{Human Baseline} \\
\midrule
Human Expert & 92.8 & 92.0 & 96.2 & 94.1 & 96.8 & 95.3 & 96.3 & 98.4 & 76.8 & 91.5 \\
\midrule
\modelgroup{Proprietary models} \\
\addlinespace[2pt]
GPT-5.6 Sol~\citep{openai2026gpt56} & 47.7 & 28.1 & 54.7 & 41.4 & 66.0 & 49.1 & 59.3 & 80.3 & 40.4 & 54.0 \\
GPT-6 Astra~\citep{openai2026gpt6astra} & \textbf{57.4} & \textbf{37.5} & \textbf{65.8} & \textbf{51.7} & \textbf{74.4} & \textbf{60.8} & \textbf{68.6} & \textbf{84.2} & \textbf{47.5} & \textbf{63.1} \\
Gemini 3.1 Pro Preview~\citep{deepmind2026gemini31pro} & 33.4 & 16.2 & 38.5 & 27.3 & 50.3 & 33.7 & 43.7 & 67.7 & 29.3 & 39.5 \\
\midrule
\modelgroup{Open-source VLMs} \\
\addlinespace[2pt]
Qwen2.5-VL-72B-Instruct~\citep{bai2025qwen25vl} & 24.3 & 7.6 & 23.8 & 15.7 & 44.1 & 26.9 & 34.7 & 51.1 & 24.9 & 32.8 \\
Qwen3-VL-32B-Instruct~\citep{bai2025qwen3vl} & 28.1 & 10.2 & 28.5 & 19.4 & 49.7 & 30.4 & 38.3 & 62.5 & 27.9 & 36.9 \\
Qwen3-VL-8B-Instruct~\citep{bai2025qwen3vl} & 23.6 & 8.7 & 24.5 & 16.6 & 42.9 & 24.4 & 32.2 & 52.6 & 22.1 & 30.6 \\
InternVL3.5-8B~\citep{wang2025internvl35} & 19.9 & 8.0 & 24.5 & 16.3 & 33.2 & 18.3 & 25.8 & 46.0 & 15.8 & 23.5 \\
MiniCPM-V-4.5-8B~\citep{openbmb2025minicpmv45} & 19.4 & 6.8 & 20.1 & 13.4 & 35.3 & 21.1 & 26.4 & 36.6 & 18.8 & 25.5 \\
VideoLLaMA3-7B~\citep{zhang2025videollama3} & 13.1 & 3.5 & 9.4 & 6.5 & 24.0 & 16.2 & 23.4 & 38.1 & 14.9 & 19.7 \\
Qwen3-VL-4B-Instruct~\citep{bai2025qwen3vl} & 24.7 & 8.4 & 25.2 & 16.8 & 46.4 & 24.4 & 35.6 & 50.2 & 22.8 & 32.5 \\
Molmo2-8B~\citep{clark2026molmo2} & 17.4 & 6.9 & 19.2 & 13.1 & 29.0 & 18.2 & 22.4 & 37.2 & 16.9 & 21.8 \\
LLaVA-OV2-8B~\citep{an2026llavaov2} & 23.9 & 8.7 & 23.8 & 16.2 & 44.0 & 24.8 & 34.2 & 48.9 & 22.8 & 31.6 \\
\midrule
\modelgroup{Open-source video captioning models} \\
\addlinespace[2pt]
video-SALMONN-2+-3B~\citep{tang2025videosalmonn2} & 10.4 & 2.9 & 12.2 & 7.6 & 20.2 & 8.9 & 14.3 & 29.2 & 9.4 & 13.3 \\
CapRL-Video-4B~\citep{yang2026caprlplusplus} & 32.0 & 15.5 & 39.0 & 27.2 & 51.0 & 28.6 & 39.6 & 65.0 & 27.0 & 36.8 \\
OmniCaptioner-IF-3B~\citep{wang2026omnicapif} & 19.9 & 6.2 & 23.9 & 15.1 & 37.6 & 18.1 & 24.7 & 28.3 & 18.8 & 24.8 \\
OmniCaptioner-IF-7B~\citep{wang2026omnicapif} & 22.8 & 9.1 & 25.5 & 17.3 & 39.2 & 23.2 & 28.5 & 43.2 & 21.5 & 28.3 \\
\midrule
\modelgroup{Open-source egocentric video understanding models} \\
\addlinespace[2pt]
EgoThinker-7B~\citep{pei2025egothinker} & 22.6 & 14.8 & 25.9 & 20.3 & 36.7 & 20.0 & 22.9 & 14.5 & 19.9 & 24.9 \\
EgoGPT-7B~\citep{yang2025egolife} & 14.5 & 3.7 & 13.4 & 8.6 & 24.2 & 18.8 & 18.1 & 22.0 & 20.6 & 20.5 \\
RynnBrain-4B~\citep{dang2026rynnbrain} & 15.1 & 10.6 & 15.2 & 12.9 & 22.9 & 13.9 & 16.9 & 17.8 & 15.8 & 17.3 \\
Flora~\citep{leng2026flora} & 34.9 & 16.9 & 43.2 & 30.0 & 48.4 & 38.4 & 34.2 & 57.9 & 37.9 & 39.9 \\
\midrule
\modelgroup{EgoAction (Ours)} \\
\addlinespace[2pt]
\textbf{EgoAction (LLaVA-OV2-8B)} & 47.5 & 30.1 & 57.5 & 43.8 & 66.7 & 45.5 & 52.2 & 64.7 & 39.7 & 51.2 \\
\textbf{EgoAction (Qwen3-VL-4B)} & 42.0 & 25.4 & 48.0 & 36.7 & 63.3 & 41.0 & 48.5 & 62.4 & 36.0 & 47.4 \\
\bottomrule
\end{tabular}
\end{adjustbox}
\end{table}

\clearpage
\begin{table}[!htbp]
\centering
\scriptsize
\setlength{\tabcolsep}{4pt}
\renewcommand{\arraystretch}{1.12}
\newcommand{\modelgroup}[1]{\multicolumn{11}{@{}l}{\textit{#1}}}
\caption{\textbf{Dense action captioning on the EgoLife subset of EgoActionBench.}
All scores range from 0 to 100; higher is better.
Temporal Overall is the mean of m-tIoU and micro-averaged F1.
Semantic Overall averages the applicable Hand, Verb, Objects, Tools, and Motion scores within each ground-truth action, then averages over ground-truth actions.
EgoActionScore is the mean of Temporal Overall and Semantic Overall.
Parentheses identify the EgoAction backbone.
Best model scores are bold.}
\label{tab:source_egolife}
\begin{adjustbox}{max width=\linewidth}
\begin{tabular}{@{}l r @{\hspace{7pt}} rrr @{\hspace{7pt}} rrrrrr@{}}
\toprule
\multirow{2}{*}{\textbf{Model}}
& \multirow{2}{*}{\makecell{\textbf{EgoAction}\\\textbf{Score}}}
& \multicolumn{3}{c}{Temporal} & \multicolumn{6}{c}{Semantic} \\
\cmidrule(lr){3-5}\cmidrule(lr){6-11}
& & m-tIoU & F1 & Overall
& Hand & Verb & Objects & Tools & Motion & Overall \\
\midrule
\modelgroup{Human Baseline} \\
\midrule
Human Expert & 93.8 & 94.4 & 97.3 & 95.9 & 97.3 & 96.5 & 96.7 & 97.1 & 76.1 & 91.8 \\
\midrule
\modelgroup{Proprietary models} \\
\addlinespace[2pt]
GPT-5.6 Sol~\citep{openai2026gpt56} & 47.4 & 29.3 & 55.1 & 42.2 & 63.6 & 48.4 & 58.1 & 72.5 & 40.0 & 52.6 \\
GPT-6 Astra~\citep{openai2026gpt6astra} & \textbf{59.0} & \textbf{40.3} & \textbf{68.6} & \textbf{54.5} & \textbf{74.3} & \textbf{63.0} & \textbf{67.8} & \textbf{77.6} & \textbf{49.1} & \textbf{63.6} \\
Gemini 3.1 Pro Preview~\citep{deepmind2026gemini31pro} & 33.0 & 16.8 & 40.1 & 28.5 & 45.1 & 33.5 & 40.7 & 58.3 & 30.5 & 37.5 \\
\midrule
\modelgroup{Open-source VLMs} \\
\addlinespace[2pt]
Qwen2.5-VL-72B-Instruct~\citep{bai2025qwen25vl} & 24.1 & 8.0 & 26.7 & 17.3 & 41.5 & 26.6 & 29.5 & 41.6 & 26.2 & 30.9 \\
Qwen3-VL-32B-Instruct~\citep{bai2025qwen3vl} & 27.2 & 11.0 & 31.5 & 21.2 & 44.4 & 28.4 & 32.6 & 42.0 & 27.5 & 33.2 \\
Qwen3-VL-8B-Instruct~\citep{bai2025qwen3vl} & 24.6 & 10.0 & 27.5 & 18.7 & 42.7 & 25.5 & 30.1 & 43.0 & 23.7 & 30.5 \\
InternVL3.5-8B~\citep{wang2025internvl35} & 19.0 & 8.1 & 23.4 & 15.8 & 29.6 & 19.5 & 21.7 & 29.0 & 18.2 & 22.2 \\
MiniCPM-V-4.5-8B~\citep{openbmb2025minicpmv45} & 18.8 & 5.9 & 19.3 & 12.6 & 32.0 & 22.7 & 24.0 & 28.7 & 21.3 & 24.9 \\
VideoLLaMA3-7B~\citep{zhang2025videollama3} & 14.8 & 4.1 & 11.0 & 7.5 & 24.5 & 19.7 & 24.3 & 33.8 & 20.1 & 22.1 \\
Qwen3-VL-4B-Instruct~\citep{bai2025qwen3vl} & 24.5 & 9.3 & 27.1 & 18.2 & 42.7 & 25.1 & 30.8 & 26.6 & 25.1 & 30.9 \\
Molmo2-8B~\citep{clark2026molmo2} & 17.1 & 6.9 & 19.4 & 13.1 & 27.4 & 17.7 & 21.8 & 31.6 & 17.1 & 21.0 \\
LLaVA-OV2-8B~\citep{an2026llavaov2} & 22.7 & 8.5 & 24.5 & 16.5 & 37.8 & 25.2 & 29.7 & 44.3 & 23.1 & 28.9 \\
\midrule
\modelgroup{Open-source video captioning models} \\
\addlinespace[2pt]
video-SALMONN-2+-3B~\citep{tang2025videosalmonn2} & 10.1 & 3.3 & 13.9 & 8.6 & 18.0 & 7.8 & 12.3 & 28.7 & 8.1 & 11.5 \\
CapRL-Video-4B~\citep{yang2026caprlplusplus} & 30.6 & 16.1 & 38.7 & 27.4 & 44.9 & 27.4 & 34.4 & 42.5 & 28.7 & 33.9 \\
OmniCaptioner-IF-3B~\citep{wang2026omnicapif} & 20.2 & 6.4 & 26.3 & 16.3 & 36.2 & 19.1 & 20.1 & 18.2 & 21.4 & 24.1 \\
OmniCaptioner-IF-7B~\citep{wang2026omnicapif} & 23.5 & 10.5 & 25.0 & 17.7 & 38.7 & 25.8 & 24.8 & 33.6 & 28.1 & 29.3 \\
\midrule
\modelgroup{Open-source egocentric video understanding models} \\
\addlinespace[2pt]
EgoThinker-7B~\citep{pei2025egothinker} & 18.9 & 12.0 & 21.9 & 16.9 & 30.8 & 19.1 & 14.2 & 6.5 & 20.0 & 20.9 \\
EgoGPT-7B~\citep{yang2025egolife} & 15.9 & 4.6 & 16.1 & 10.3 & 24.8 & 20.4 & 19.2 & 18.4 & 21.3 & 21.4 \\
RynnBrain-4B~\citep{dang2026rynnbrain} & 13.3 & 10.3 & 11.8 & 11.0 & 21.1 & 12.8 & 14.5 & 9.9 & 14.2 & 15.6 \\
Flora~\citep{leng2026flora} & 36.1 & 16.1 & 43.7 & 29.9 & 49.5 & 41.3 & 37.2 & 59.9 & 41.0 & 42.2 \\
\midrule
\modelgroup{EgoAction (Ours)} \\
\addlinespace[2pt]
\textbf{EgoAction (LLaVA-OV2-8B)} & 46.3 & 31.6 & 55.2 & 43.4 & 64.4 & 45.1 & 48.7 & 53.9 & 38.8 & 49.3 \\
\textbf{EgoAction (Qwen3-VL-4B)} & 40.2 & 25.8 & 45.6 & 35.7 & 60.5 & 40.1 & 42.9 & 43.0 & 35.8 & 44.8 \\
\bottomrule
\end{tabular}
\end{adjustbox}
\end{table}

\clearpage
\begin{table}[!htbp]
\centering
\scriptsize
\setlength{\tabcolsep}{4pt}
\renewcommand{\arraystretch}{1.12}
\newcommand{\modelgroup}[1]{\multicolumn{11}{@{}l}{\textit{#1}}}
\caption{\textbf{Dense action captioning on the EPIC-KITCHENS-100 subset of EgoActionBench.}
All scores range from 0 to 100; higher is better.
Temporal Overall is the mean of m-tIoU and micro-averaged F1.
Semantic Overall averages the applicable Hand, Verb, Objects, Tools, and Motion scores within each ground-truth action, then averages over ground-truth actions.
EgoActionScore is the mean of Temporal Overall and Semantic Overall.
Parentheses identify the EgoAction backbone.
Best model scores are bold.}
\label{tab:source_epic_kitchens_100}
\begin{adjustbox}{max width=\linewidth}
\begin{tabular}{@{}l r @{\hspace{7pt}} rrr @{\hspace{7pt}} rrrrrr@{}}
\toprule
\multirow{2}{*}{\textbf{Model}}
& \multirow{2}{*}{\makecell{\textbf{EgoAction}\\\textbf{Score}}}
& \multicolumn{3}{c}{Temporal} & \multicolumn{6}{c}{Semantic} \\
\cmidrule(lr){3-5}\cmidrule(lr){6-11}
& & m-tIoU & F1 & Overall
& Hand & Verb & Objects & Tools & Motion & Overall \\
\midrule
\modelgroup{Human Baseline} \\
\midrule
Human Expert & 83.0 & 75.9 & 84.2 & 80.1 & 91.4 & 91.1 & 91.2 & 88.6 & 69.3 & 85.9 \\
\midrule
\modelgroup{Proprietary models} \\
\addlinespace[2pt]
GPT-5.6 Sol~\citep{openai2026gpt56} & 55.6 & 37.6 & 64.5 & 51.0 & 73.0 & 56.4 & 65.3 & 76.0 & 45.1 & 60.1 \\
GPT-6 Astra~\citep{openai2026gpt6astra} & \textbf{63.6} & \textbf{45.5} & \textbf{72.3} & \textbf{58.9} & \textbf{79.7} & \textbf{67.4} & \textbf{74.5} & \textbf{77.9} & \textbf{50.9} & \textbf{68.3} \\
Gemini 3.1 Pro Preview~\citep{deepmind2026gemini31pro} & 39.1 & 22.4 & 48.2 & 35.3 & 53.4 & 36.9 & 47.1 & 63.4 & 33.0 & 42.8 \\
\midrule
\modelgroup{Open-source VLMs} \\
\addlinespace[2pt]
Qwen2.5-VL-72B-Instruct~\citep{bai2025qwen25vl} & 25.9 & 9.5 & 28.7 & 19.1 & 44.9 & 26.8 & 33.9 & 50.9 & 25.1 & 32.7 \\
Qwen3-VL-32B-Instruct~\citep{bai2025qwen3vl} & 33.2 & 15.1 & 37.4 & 26.3 & 51.6 & 34.3 & 41.2 & 60.9 & 33.2 & 40.2 \\
Qwen3-VL-8B-Instruct~\citep{bai2025qwen3vl} & 28.3 & 13.1 & 34.1 & 23.6 & 46.1 & 27.2 & 34.6 & 54.5 & 23.4 & 33.0 \\
InternVL3.5-8B~\citep{wang2025internvl35} & 22.9 & 10.9 & 29.1 & 20.0 & 34.8 & 21.1 & 29.6 & 50.8 & 17.0 & 25.9 \\
MiniCPM-V-4.5-8B~\citep{openbmb2025minicpmv45} & 20.1 & 7.5 & 21.6 & 14.5 & 34.1 & 22.1 & 24.9 & 34.5 & 21.5 & 25.6 \\
VideoLLaMA3-7B~\citep{zhang2025videollama3} & 15.3 & 6.0 & 13.9 & 9.9 & 23.8 & 16.6 & 27.6 & 41.8 & 14.0 & 20.6 \\
Qwen3-VL-4B-Instruct~\citep{bai2025qwen3vl} & 29.3 & 12.9 & 33.7 & 23.3 & 50.6 & 26.3 & 39.3 & 55.5 & 24.7 & 35.4 \\
Molmo2-8B~\citep{clark2026molmo2} & 22.9 & 10.9 & 26.1 & 18.5 & 36.4 & 22.3 & 31.4 & 51.7 & 17.9 & 27.2 \\
LLaVA-OV2-8B~\citep{an2026llavaov2} & 28.2 & 13.0 & 33.2 & 23.1 & 44.2 & 28.1 & 33.7 & 49.0 & 26.8 & 33.3 \\
\midrule
\modelgroup{Open-source video captioning models} \\
\addlinespace[2pt]
video-SALMONN-2+-3B~\citep{tang2025videosalmonn2} & 12.4 & 4.5 & 17.5 & 11.0 & 22.0 & 8.6 & 14.6 & 27.2 & 9.3 & 13.7 \\
CapRL-Video-4B~\citep{yang2026caprlplusplus} & 34.9 & 19.8 & 43.2 & 31.5 & 51.9 & 30.5 & 43.5 & 64.1 & 26.9 & 38.4 \\
OmniCaptioner-IF-3B~\citep{wang2026omnicapif} & 19.7 & 7.3 & 27.6 & 17.5 & 33.6 & 16.4 & 19.6 & 35.0 & 18.0 & 21.9 \\
OmniCaptioner-IF-7B~\citep{wang2026omnicapif} & 22.8 & 11.3 & 24.4 & 17.8 & 37.2 & 24.3 & 24.7 & 41.2 & 25.2 & 27.9 \\
\midrule
\modelgroup{Open-source egocentric video understanding models} \\
\addlinespace[2pt]
EgoThinker-7B~\citep{pei2025egothinker} & 20.4 & 14.2 & 24.0 & 19.1 & 32.7 & 17.3 & 19.1 & 14.6 & 18.1 & 21.7 \\
EgoGPT-7B~\citep{yang2025egolife} & 16.0 & 5.8 & 19.7 & 12.7 & 23.2 & 17.3 & 17.7 & 25.4 & 18.6 & 19.2 \\
RynnBrain-4B~\citep{dang2026rynnbrain} & 12.8 & 10.1 & 12.8 & 11.4 & 18.7 & 11.5 & 14.1 & 15.7 & 12.4 & 14.1 \\
Flora~\citep{leng2026flora}  & 44.0 & 24.5 & 55.3 & 39.9 & 57.5 & 46.2 & 42.1 & 60.1 & 46.7 & 48.2 \\

\midrule
\modelgroup{EgoAction (Ours)} \\
\addlinespace[2pt]
\textbf{EgoAction (LLaVA-OV2-8B)} & 50.5 & 36.4 & 58.3 & 47.4 & 70.4 & 48.5 & 54.6 & 65.4 & 40.6 & 53.6 \\
\textbf{EgoAction (Qwen3-VL-4B)} & 43.6 & 29.4 & 43.9 & 36.6 & 67.1 & 43.9 & 52.0 & 60.2 & 39.2 & 50.6 \\
\bottomrule
\end{tabular}
\end{adjustbox}
\end{table}

\FloatBarrier

\subsection*{AI Use Statement}

We used GPT-6 Astra to assist with benchmark annotation. Human reviewers
verified the benchmark annotations against the original videos, whereas
the labels used for supervised adaptation were generated and refined
automatically. Qwen3-32B served as the text-only semantic judge described in
our evaluation protocol. AI tools also assisted with code development and
language polishing. We take full
responsibility for the final paper, dataset annotations, code, and reported
results.

\end{document}